%% file: MAIN_AND_SUPP.tex
\documentclass[runningheads]{llncs}

\usepackage{eccv}

\usepackage{eccvabbrv}

\usepackage{graphicx}
\usepackage{booktabs}

\usepackage[accsupp]{axessibility}  % Improves PDF readability for those with disabilities.

\input{preamble}

\usepackage{xr}
\usepackage{xr-hyper}
\usepackage[pagebackref,breaklinks,colorlinks,citecolor=eccvblue]{hyperref}

\usepackage{orcidlink}

\begin{document}

\title{\ourtitle} 

\titlerunning{\ourtitle}

% Second Author\inst{2,3}\orcidlink{1111-2222-3333-4444} \and
% Third Author\inst{3}\orcidlink{2222--3333-4444-5555}}

\author{
Hila Manor\inst{2\textsuperscript{\dag},1}\orcidlink{0009-0007-6851-148X} \and
Rinon Gal\inst{2\textsuperscript{\dag}}\orcidlink{0000-0003-4875-965X} \and
Haggai Maron\inst{2,1}\orcidlink{0009-0001-4088-6286} \and
Tomer Michaeli\inst{1}\orcidlink{0000-0003-0525-8054} \and
Gal Chechik\inst{2,3}\orcidlink{0000-0001-9164-5303}
}

\authorrunning{H.~Manor et al.}

\institute{
$^1$Technion, Israel $\quad$ $^2$NVIDIA, Israel $\quad$ $^3$Bar-Ilan University, Israel
% \\ \dag This work was done while HM and RG were working at NVIDIA.
% Technion, Israel \and NVIDIA Research, Israel \and Bar-Ilan University, Israel
}

% Springer Heidelberg, Tiergartenstr.~17, 69121 Heidelberg, Germany

% \maketitle

\begingroup
\renewcommand{\thefootnote}{\dag}
\footnotetext[0]{
\small
This work was done while HM and RG were working at NVIDIA.}
\endgroup

% \twocolumn[{%
% \renewcommand\twocolumn[1][]{#1}%
\maketitle
\begin{center}
    \centering
    \captionsetup{type=figure}
    \includegraphics[width=0.989\linewidth]{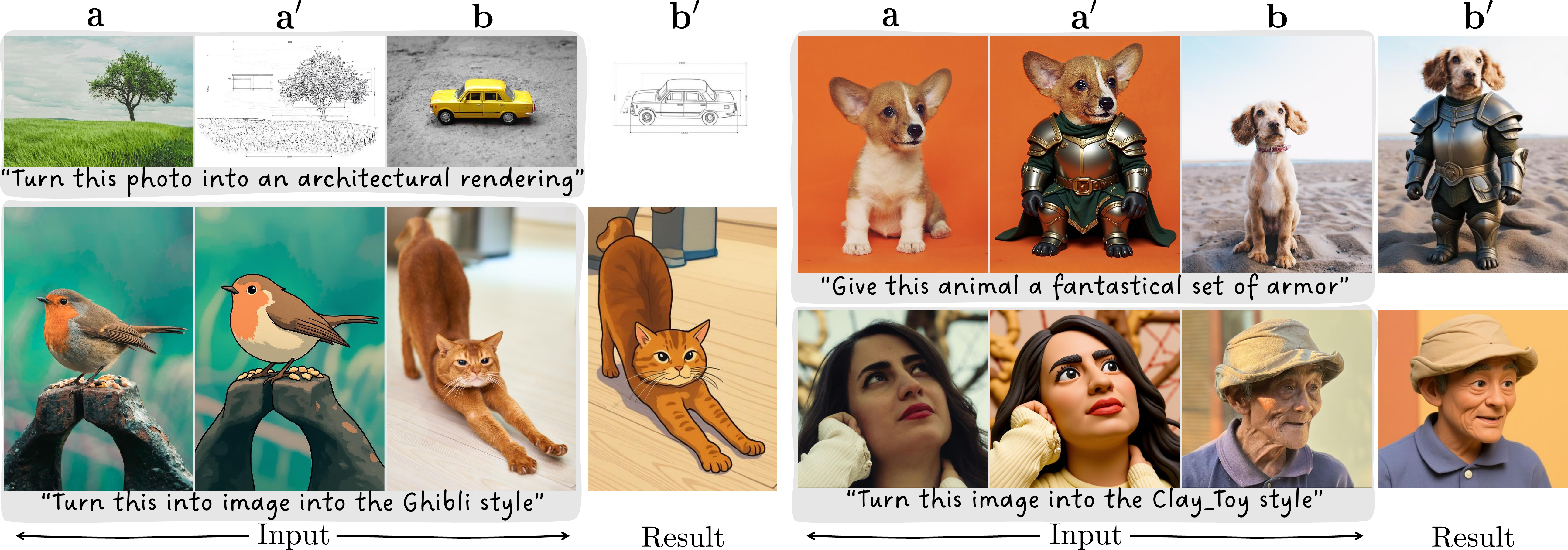}
    \captionof{figure}{
    \textbf{\coolname.} 
    We present a novel method for analogy-based editing via learnable mixing of LoRAs.
    Given a prompt and an image triplet $\{\rva,\rva',\rvb\}$ depicting a desired transformation, \coolname dynamically constructs a single LoRA from a learnable basis of LoRAs, and produces an editing result $\rvb'$ that applies the same analogy to $\rvb$. 
    }
    \label{fig:teaser}
\end{center}%
% }]

\input{sec/0_abstract}    
\input{sec/1_intro}

\input{sec/2_related_work}
\input{sec/3_method}

\input{sec/4_experiments}
\input{sec/5_discussion}

\section*{Acknowledgments}
This research was partially supported by the Israel Science
Foundation (grant no. 2318/22) and the Planning and Budgeting Committee of the Israeli Council for Higher Education. 
The authors are grateful to Matan Kleiner and Yoad Tewel for their insightful discussions and input throughout this work.

\bibliographystyle{splncs04}
\bibliography{main}

\beginsupplement
\appendix

% Tell cleveref to treat sections and subsections here as appendices
\crefalias{section}{appendix}
\crefalias{subsection}{appendix}
\crefalias{subsubsection}{appendix}

% \maketitlesupplementary
\onecolumn
% \maketitle

\input{sec/app_implementation_details}

\clearpage
\input{sec/app_more_results}

\end{document}

%% file: preamble.tex
\usepackage{makecell}
\usepackage{multirow}
\usepackage{placeins}
\usepackage{tabularx}

\usepackage{listings}
\makeatletter
\@ifundefined{citep}{ \newcommand{\citep}{\cite} }{}
\@ifundefined{citet}{ \newcommand{\citet}{\cite} }{}
\makeatother

\newcommand{\ourtitle}{Spanning the Visual Analogy Space with a Weight Basis of LoRAs}

\newcommand{\coolname}{LoRWeB\xspace}
\newcommand{\ourmethod}{LoRWeB\xspace}
\newcommand{\flux}{Flux.1-Kontext\xspace}

\def\naive{na\"{\i}ve\xspace}

\input{math_commands}

\makeatletter
\DeclareRobustCommand\onedot{\futurelet\@let@token\@onedot}
\def\@onedot{\ifx\@let@token.\else.\null\fi\xspace}
\def\eg{\emph{e.g}\onedot}

\def\ie{\emph{i.e}\onedot}

\makeatother

\makeatletter
\newcommand*{\addFileDependency}[1]{% argument=file name and extension
  \typeout{(#1)}
  \@addtofilelist{#1}
  \IfFileExists{#1}{\typeout{File #1 O.K.}}{\typeout{No file #1.}}
}
\makeatother

\newcommand{\beginsupplement}{%
    \renewcommand\thefigure{S\arabic{figure}}    
    \setcounter{figure}{0}
    \renewcommand\thetable{S\arabic{table}}    
    \setcounter{table}{0}  
    \renewcommand\theequation{S\arabic{equation}}    
    \setcounter{equation}{0}
}

\AtEndPreamble{
    \usepackage[capitalize]{cleveref}
\crefname{appendix}{App.}{Apps.}
\Crefname{appendix}{Appendix}{Appendices}
}

%% file: math_commands.tex
\usepackage{amsmath,amsfonts,bm}

\def\eqref#1{eq.~(\ref{#1})}
\def\1{\bm{1}}

\def\rva{{\mathbf{a}}}
\def\rvb{{\mathbf{b}}}

\def\rvk{{\mathbf{k}}}

\def\rvq{{\mathbf{q}}}

\def\rvx{{\mathbf{x}}}
\def\rvy{{\mathbf{y}}}
\def\rvz{{\mathbf{z}}}

\def\mA{{\bm{A}}}
\def\mB{{\bm{B}}}

\def\mK{{\bm{K}}}

\def\mW{{\bm{W}}}

\DeclareMathAlphabet{\mathsfit}{\encodingdefault}{\sfdefault}{m}{sl}
\SetMathAlphabet{\mathsfit}{bold}{\encodingdefault}{\sfdefault}{bx}{n}

\def\gE{{\mathcal{E}}}

\def\gL{{\mathcal{L}}}

\def\gP{{\mathcal{P}}}

\def\gT{{\mathcal{T}}}

\def\sR{{\mathbb{R}}}

\newcommand{\E}{\mathbb{E}}

%% file: sec/0_abstract.tex
\begin{abstract}
Visual analogy learning enables image editing via demonstration rather than textual description, allowing users to specify complex transformations difficult to articulate in words. Given a triplet $\{\rva$, $\rva'$, $\rvb\}$, the goal is to generate $\rvb'$ such that $\rva\!:\!\rva'\!::\!\rvb\!:\!\rvb'$. 
Recent methods adapt text-to-image models 
with a single Low-Rank Adaptation (LoRA) module, but they face 
a fundamental limitation: attempting to capture the diverse space of visual transformations within a fixed 
module constrains generalization. 
Inspired by recent work showing that LoRAs in constrained domains span meaningful, interpolatable semantic spaces,
we propose \coolname, 
which specializes the model for each analogy task in a single inference pass. \coolname dynamically composes learned transformation primitives, informally, choosing a point in a ``\emph{space of LoRAs}''. We introduce two key components: (1) a learnable basis of LoRAs to span the space of different visual transformations,  and (2) a lightweight encoder that dynamically weighs these basis LoRAs given the input analogy pair. 
Comprehensive evaluations demonstrate state-of-the-art performance and significantly improved generalization to unseen transformations. Our findings suggest LoRA basis decompositions are a promising direction for flexible visual manipulation tasks.
 See our \href{https://research.nvidia.com/labs/par/lorweb}{website} for code.
\keywords{Image analogies \and Image editing \and LoRA \and Flow models}
\end{abstract}

%% file: sec/1_intro.tex
\section{Introduction}

Text-based image editing models~\citep{labs2025kontext,brooks2023instructpix2pix,xiao2025omnigen,sheynin2024emu,zhang2025context} have recently emerged as powerful tools for controllable image generation and manipulation, enabling users to modify images through textual descriptions. However, many visual transformations are inherently difficult to articulate precisely through text alone. For example, consider describing the transformation that converts a photograph into the style of a specific painting, or conveying an exact target pose through text. Such inherent limitations motivate the need for alternative paradigms that can capture and apply complex visual transformations.

Visual analogy learning~\citep{imageanalogies} offers a compelling solution to this challenge by enabling models to understand transformations through examples rather than explicit descriptions. In this paradigm, given a triplet of images $\{\rva, \rva', \rvb\}$, the goal is to generate an image $\rvb'$ such that the visual relationship $\rva:\rva' :: \rvb:\rvb'$ holds. That is, the transformation applied between $\rva$ and $\rva'$ should be analogously applied to $\rvb$ to produce $\rvb'$. This approach allows users to specify complex visual changes through demonstration, making it possible to capture nuanced transformations that would be difficult or impossible to describe textually.

Early learning-based approaches trained stand-alone analogy models directly from analogy data~\citep{reed2015deep,bar2022visual,wang2023images,liu2024unifying,wang2023context,yang2023imagebrush}, but this lead to limited task diversity and image quality, or required extensive compute. More recent work aims to leverage the rich prior of powerful text-to-image backbones by adapting them to the visual analogy task, using a single Low-Rank Adaptation (LoRA) module~\citep{lu2025pairedit,song2024lora,gong2025relationadapter}. While effective, these methods face a fundamental limitation: they attempt to capture the diverse space of possible transformations within a single adaptation module. This constraint may limit the model's ability to generalize across the rich variety of relationships that exist in images.

We hypothesize that specializing the model to each specific analogy task at inference time may improve performance and generalization. While this objective could theoretically be achieved via hypernetworks that generate task-specific LoRAs~\citep{song2024lora}, these are notoriously difficult to train and often suffer from instability~\citep{ortiz2024magnitude}. 
Instead, we draw inspiration from recent work~\citep{dravid2024interpreting} demonstrating that fine-tuned LoRAs (\eg, for personalization tasks) can form an interpretable weight space. In this space, individually-trained LoRAs act as a basis of fundamental and semantic visual traits, and interpolating between the weights of these LoRAs can effectively cover new points in this semantic space, which allows for creating new, blended concepts.
Building on this insight, we explore a similar principle for visual analogy learning and propose \coolname, a two-component system: (1) a learnable basis of LoRA modules and (2) a lightweight encoder that dynamically combines LoRAs from this basis at inference time based on the input analogy pair. These components are jointly trained, enabling the model to compose appropriate transformations for novel analogies unseen during training.

Existing methods typically encode analogy images using vision-language models such as CLIP~\citep{clip} or SigLIP~\citep{zhai2023sigmoid} and provide these encodings as context to the generative model.
This can provide the higher-level semantic understanding needed for understanding the analogy task. However, this might lead to loss of detail in fine-grained visual detail preservation.
Recent advances have shown that diffusion models can extract remarkably accurate visual details through extended attention mechanisms~\citep{cao2023masactrl,labs2025kontext}. Thus, we leverage this capability by providing the full analogy triplet directly to the diffusion model through an extended-attention mechanism, while reserving CLIP-based encodings specifically for LoRA selection. This approach allows \coolname to balance fine-detail consistency with the higher-level semantics required to understand the analogy task. 

We evaluate \coolname against established baselines and show it achieves state-of-the-art results. Our contributions include: (1) a novel architecture that decomposes visual analogy learning into a basis of LoRAs with dynamic composition, and (2) a comprehensive evaluation showing improved generalization to unseen transformations compared to existing single-LoRA approaches.

%% file: sec/2_related_work.tex
\section{Related Work}

\subsubsection{Visual Analogies.}
Visual analogies, also known as ``Image Analogies''~\citep{imageanalogies}, ``Visual Prompting''~\citep{bar2022visual} or ``Visual Relations''~\citep{gong2025relationadapter}, is the task of learning a transformation from a pair of before-and-after exemplars and applying it analogously to new images. 
Early non-neural methods learned explicit per-pair filters for simpler tasks such as style transfer~\citep{imageanalogies}, or per-pair optimization for relighting in 3D~\cite{fivser2016stylit}. 
With the advent of network-based methods, initial works proposed models conditioned on 
image embeddings or NeRF~\cite{mildenhall2021nerf} representations to present analogies through simple vector arithmetic~\citep{reed2015deep,liao2017visual,he2019progressive,fischer2024nerf}. While these methods showed promise on datasets of simple, isolated objects, they struggled with the complexity of real-world images, and still mostly tackled style-transfer analogies.
Newer methods instead treat analogy learning as in-context learning, where the model is directly conditioned on the exemplar pair and a reference image, and is trained to successfully synthesize the matching target~\citep{bar2022visual,wang2023context,wang2023images,yang2023imagebrush}.
More recently, some works adapt pre-trained text-to-image foundation models to the new task, going beyond simple style-transfer analogies.
For example, \cite{vsubrtova2023diffusion} use per-sample optimization, backpropegating through the entire diffusion process. However, such approaches can require dozens of minutes to edit every image, and their memory requirements can be challenging with newer, larger models. 
Another approach adapts the foundation model directly using a LoRA module~\citep{hu2022lora,gong2025relationadapter,chen2025edit}.
These methods, while showing impressive results, still struggle to generalize to unseen tasks. Our approach aims to tackle this limitation by avoiding the bottleneck of a single LoRA, opting instead to train a basis of adapters which can be mixed to achieve greater flexibility and better generalization.

\subsubsection{Diffusion-Based Image Editing.}

The unprecedented semantic control offered by large scale text-to-image diffusion models~\citep{rombach2021highresolutionLDM,ramesh2022hierarchical,labs2025kontext} has inspired extensive work leveraging them as priors for image editing.
Early works add noise to an image and remove it conditioned on a novel prompt~\citep{meng2022sdedit}, though such methods significantly change image structure. 
Subsequent work improved content preservation by manipulating internal feature representations~\citep{hertz2022prompt,parmar2023zeroshot,tumanyan2022plug} or the model's denoising trajectory~\citep{hertz2023delta,kulikov2024flowedit,deutch2024turboedit,HubermanSpiegelglas2023}.
Recent works go beyond text and incorporate different control modalities for enhanced precision, such as ControlNet~\citep{cao2023masactrl, zhang2023adding}, or attention-sharing~\citep{tewel2024trainingfree,hertz2023style,alaluf2023cross,gal2024lcm}.
Others explore text-free editing to enable modifications that cannot be textually described~\citep{haas2024discovering,pmlr-v235-manor24a}, though without direct control.
Transformer-based diffusion models further popularized attention-sharing for maintaining subject consistency in personalization~\citep{gal2022image,ruiz2022dreambooth} and editing~\citep{tewel2024add,tan2025ominicontrol,cai2025diffusion}.
Among these, \flux~\citep{labs2025kontext} was specifically trained for text-based editing, incorporating input images via extended attention mechanisms. Our work extends this model's capabilities to visual analogies.

\subsubsection{LoRA and Weight Bases.}

LoRA~\citep{hu2022lora} is a parameter-efficient fine-tuning method that modifies a model using low-rank matrices learned on top of the existing weights. Its success lead to a range of downstream approaches trying to improve on the original formula. Of these, a line of work explores the combination of multiple LoRa modules, either to combine them post-tuning~\citep{shah2024ziplora,zhang2025subject}, or as a means of turning an existing model into a mixture of experts~\citep{feng2024mixture,wu2024mixture,mao2025omni}.
In visual content generation, a recent work~\citep{dravid2024interpreting} showed that independently trained LoRA weights can span a semantic basis, and interpolations between them can be meaningful.
However, to use their findings on a practical task such as face personalization, their approach required 65,000 independently trained LoRAs, and further employed PCA to span this basis, using test-time tuning per subject. 
Similar observations on weight bases of LoRAs were made in language processing:
LoraHub~\citep{huang2024lorahub} independently trained LoRAs and combined them post-tuning with test-time optimization to enable new tasks at inference given multiple reference outputs of the new tasks, and Sci-LoRA~\citep{cheng2025sci} combined LoRAs
for tasks like text simplification across different scientific domains. 
We propose to further expand on this idea by learning a joint basis of LoRAs, along with the router to mix and match between them efficiently at inference time. 
Thus, we can learn a basis that is more amenable to interpolations, and enable better downstream generalization.
Our approach does not necessitate the extensive inference-time compute required by prior approaches, rather creating an input dependent LoRA with a single forward pass at inference time, without test-time optimization.

%% file: sec/3_method.tex
\section{Method}

\begin{figure*}[t]
    \centering
    \includegraphics[width=\linewidth]{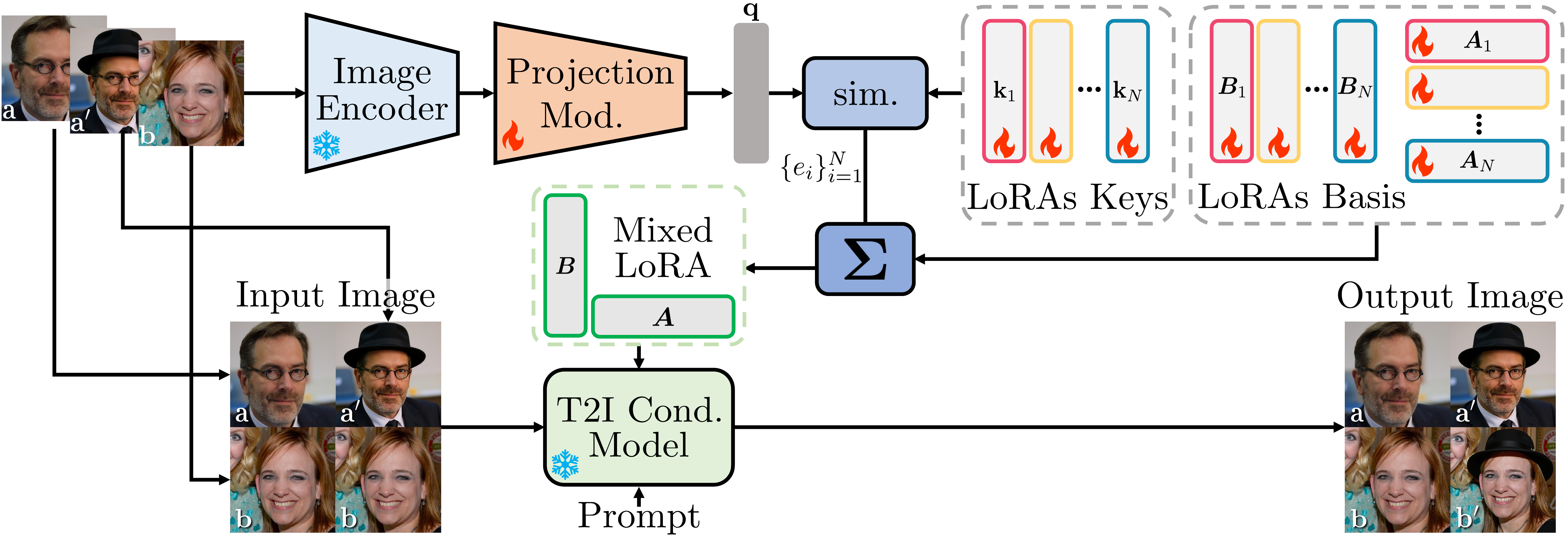}
    \caption{\textbf{\coolname Overview.}
    We first encode $\rva$ and $\rva'$, that describe a visual transformation (\eg adding a hat to the man), and $\rvb$, which should be edited analogously (\eg adding a hat to the woman) with CLIP~\citep{clip}, and a small learned projection module.  
    The similarity between the encoded vector and a set of learned keys determines the linear coefficients for combining the learned LoRAs into a single, mixed LoRA. This mixed LoRA is injected into a conditional flow model (\eg Flux.1-Kontext~\citep{labs2025kontext}).
    Next, we build a $2\times2$ composite image from $\{\rva,\rva',\rvb\}$. The conditional flow model gets this composite image as its input, along with a guiding edit prompt, and produces a composite image with the edited results $\rvb'$ in the bottom-right quadrant.}
    \label{fig:overview}
\end{figure*}

\subsection{Preliminaries}

\subsubsection{Low-Rank Adaption.}
LoRA~\citep{hu2022lora} offers a parameter-efficient alternative to conventional fine-tuning of large models by learning low-rank matrices that adapt the pre-trained weights. Specifically, starting from a frozen pre-trained weight matrix $\mW_0\in\sR^{m\times n}$, the update of the weights is represented as the product of two learned low-rank matrices $\Delta \mW=\mB\mA$, where $\mB \in \sR^{m \times r}$ and $\mA \in \sR^{r \times n}$, and the rank $r$ is typically $r \ll \min(m, n)$. This formulation drastically reduces the number of trainable parameters, while typically maintaining model performance.
The final weights of the model are then updated to $\smash{\mW = \mW_0 + \frac{\alpha}{r}\mB\mA}$, where $\alpha$ is a scaling constant.

\subsubsection{Flow Models.}
Flow-based generative models~\citep{liu2023flow,lipman2023flow,albergo2023building} learn a series of transformations to map samples from one probability distribution $\rvx_1\sim p$, to samples from another $\rvx_0\sim q$. In the generative context, $p$ is typically taken as the standard normal distribution, while $q$ is the data distribution in a latent space~\citep{rombach2021highresolutionLDM}. 
Then, These models learn a time-dependent velocity field $v_\theta(\rvz_t, t)$ that models the direction from a noisy sample towards the data manifold. The noisy sample $\rvz_t$ is a linearly interpolated latent between the two data distributions, and is given as $\rvz_t = (1 - t)\rvx_0 + t\rvx_1$. The rectified flow-matching training loss for a conditional model conditioned on a text prompt $c$ is given by 
\begin{align}\label{eq:fluxloss}
\gL=\E_{t \sim p(t),\rvx_0, \rvx_1, \rvy, c} \left[ \left\| v_\theta(\rvz_t, t, \rvy, c) - (\rvx_1-\rvx_0) \right\|_2^2 \right].
\end{align}
Here, the velocity field is optionally conditioned on a context image $\rvy$.

\subsection{\coolname}
Our objective is to perform visual analogy completion~\citep{imageanalogies}, where the model infers a proposed edit from a given image pair and applies it to a new image. Formally, two reference images, $\rva,\rva'\in\sR^D$, are related by an unknown transformation $\gT:\sR^D\rightarrow \sR^D$ such that $\rva'=\gT(\rva)$. Given a new image $\rvb\in\sR^D$, the goal is to generate $\rvb'\in\sR^D$ such that $\rvb'\approx\gT(\rvb)$. 

\subsubsection{Naive Solutions and Limitations.} Using a pre-trained conditional generative model, such as FLUX.1-Kontext~\citep{labs2025kontext}, existing solutions for this task fine-tune the model using a single LoRA~\citep{ryu2023low}. For example, given the input triplet $\{\rva,\rva',\rvb\}$, one can construct a composite $2\times2$ image $\rvy=\left[\rva, \rva' ; \rvb, \rvb \right]$, as shown in the bottom-left part of \cref{fig:overview}, which serves as the conditioning input. The goal of the model is to output $\rvx_0=\left[\rva, \rva' ; \rvb, \rvb' \right]$, such that the bottom-right quadrant was transformed from $\rvb$ to $\rvb'$, by training over \cref{eq:fluxloss}.
While these approaches perform well when the transformation $\gT$ is constrained to the analogy types seen in the training set, they struggle to generalize to new, diverse transformations. 
We propose this arises in part because the single adapter struggles to capture the wide range of analogical relationships, from different style transfers to objects insertion or layout modifications. 

A more advanced solution could be to span the diverse set of possible analogies using multiple adapters. Recently, \citet{dravid2024interpreting} demonstrated that LoRAs trained for model personalization can span a semantic basis. Inspired by this, we propose to learn such a basis for \emph{task LoRAs}. 
A \naive adaptation of \citet{dravid2024interpreting} to analogy tasks would require us to first optimize a single adapter for each of $N$ analogy types seen during training, such that each LoRA module $i$ excels at a different subset of visual edits. Once the specialized adapters are trained, they can be linearly combined to obtain an equivalent single ``novel'' adapter 
\begin{align}\label{eq:linear_comb}
    \mA=\sum e_i \mA_i, \quad \mB=\sum e_i \mB_i,
\end{align}
where the coefficients $e_i$ are optimized for each analogy task separately through the use of \cref{eq:fluxloss} and the reference pair of images $\{\rva,\rva'\}$. The model using the combined LoRA is then used to transform $\rvb$ to $\rvb'$. 

However, this approach requires training a large number of models, and a test-time tuning phase for every new analogy. Indeed, \citet{dravid2024interpreting} required $65,000$ LoRAs to capture the constrained space of faces, and collecting a significant number of different analogy pairs is more difficult. 

\subsubsection{Our Approach.} Instead, we propose \coolname (\textbf{Lo}w-\textbf{R}ank \textbf{We}ight \textbf{B}asis). Rather than training individual LoRAs and combining them only at inference time, we propose to simultaneously train a basis of LoRA adapters, jointly with an encoder that predicts linear-combination coefficients for each input analogy pair.
Specifically, we maintain a set of $N$ rank-$r$ LoRAs, and associate each $\mA_i,\mB_i$ pair where $i\in\{1,\ldots,N\}$ with a learnable key vector $\rvk_i \in \sR^d$, as depicted in the right part of \cref{fig:overview}. 
Next, we define an encoder network based on a frozen, pre-trained ViT~\citep{zhai2022scaling}, $\gE$, \eg CLIP~\citep{clip}. 
The encoder takes as input the conditioning image triplet, $\{\rva,\rva',\rvb\}$, passes them through the ViT, concatenates the results and projects them through a small learnable projection module $\gP$ that outputs the results as a query vector $\rvq\in\sR^d$:
\begin{align}
    \rvq(\rva,\rva',\rvb) = \gP\Big(\big[\gE(\rva), \gE(\rva'), \gE(\rvb)\big]\Big).
\end{align}
Then, based on the conditioning query, we compute $N$ coefficients with 
\begin{align}\label{eq:sim}
    e_i(\rva,\rva',\rvb) = \left[\text{softmax}\left( \frac{\rvq(\rva,\rva',\rvb)\mK^T}{\sqrt{d}} \right) \right]_i,
\end{align}
where $K\in\sR^{d\times N}$ contains the key vectors $\{\rvk_i\}_{i=1}^N$ in its columns. The final LoRA combination follows 
\begin{align}\label{eq:our_comb}
    \Delta \mW=\mB\mA=\sum e_i (\mB_i \mA_i ),
\end{align}
and is marked as ``Mixed LoRA'' in \cref{fig:overview}.

Importantly, a single rank-$r'$ LoRA is a special case of \coolname.
Specifically, if the learned mixing router collapses to a constant output for any input $\{\rva,\rva'\rvb\}$, then $e_i$ will be constant for all inputs. This will result in the same static mixed $N$ rank $r$ LoRA combination. This mixture can yield a matrix of rank  $r'=Nr$, which is equivalent to finetuning a single rank $r'$ LoRA.
However, as \coolname differently combines LoRAs for different inputs, this dramatically increases the expressive power of our representation. Indeed, a single rank $r'$ LoRA is just a single point in the space of all rank $r'$ LoRAs, which \coolname aims to span.  

We use the same pre-trained encoder across different network layers, but train individual \coolname modules, including LoRAs, keys and projections for each targeted weight matrix $\mW_0$ in the network.
This enables capturing different semantic elements for each weight and layer in the model.

%% file: sec/4_experiments.tex
\section{Experiments}\label{sec:experiments}

\begin{figure*}[t]
    \centering
    \includegraphics[width=\linewidth]{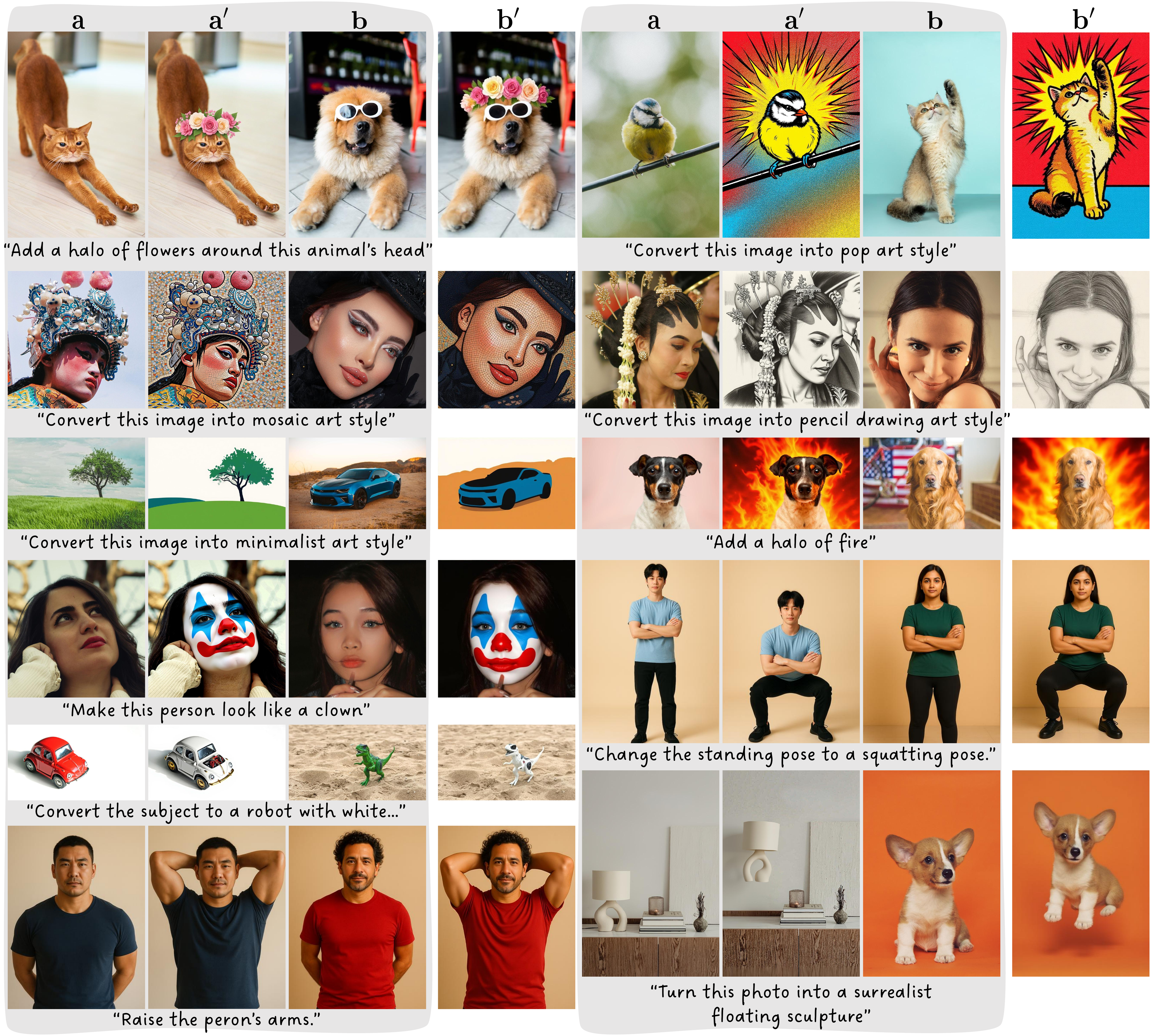}
    \caption{\textbf{\coolname visual analogy results.} Using a LoRA Basis allows \coolname to generalize to a wide variety of new analogy tasks, from adding objects to transferring specific styles or makeup or copying pose changes. Please zoom in for more details. 
    }
    \label{fig:our_res}
\end{figure*}

\begin{figure*}[t]
    \centering
    \includegraphics[width=0.99\linewidth]{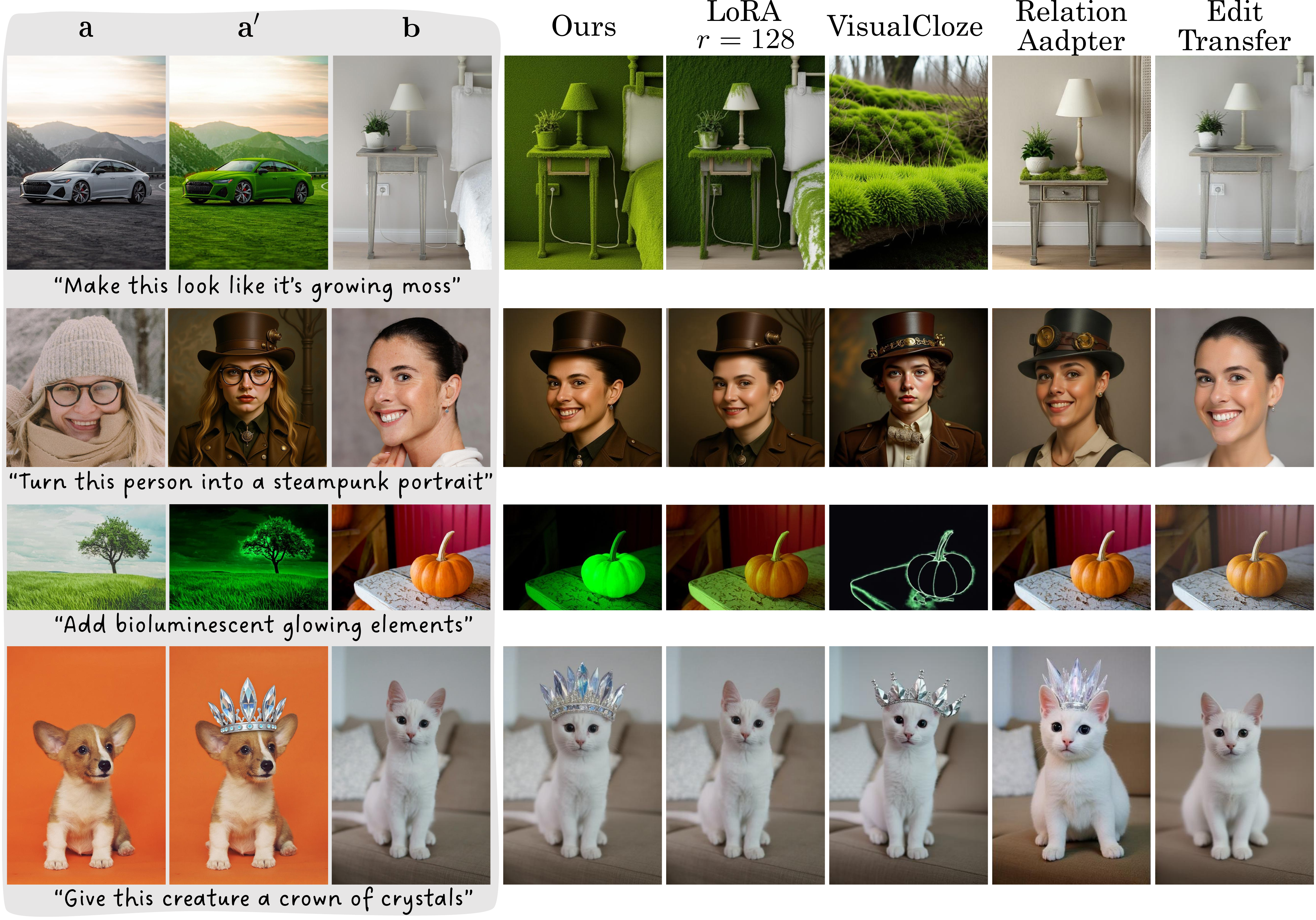}
    \caption{\textbf{Comparisons with baseline methods on unseen tasks.} 
    We compare \coolname with four recent baselines: RelationAdapter~\citep{gong2025relationadapter}, VisualCloze~\citep{li2025visualcloze} and EditTransfer~\citep{chen2025edit}, as well as a standard \flux LoRA of similar parameter capacity.
    Our approach generalizes across more diverse tasks, and better maintains the visual details of both the subject and the analogy. 
    }
    \label{fig:comparisons}
\end{figure*}

\subsubsection{Settings.}
We evaluate our approach using Flux.1-Kontext~\citep{labs2025kontext} as the pre-trained conditional flow model and CLIP~\citep{clip} as the image encoder backbone. 
For our LoRAs Basis, we match the capacity of prior work~\citep{gong2025relationadapter}, using $N=32$ adapters, each of rank $r=4$, with $d=128$ as the learned key dimension. 
We project the CLIP-encoder's output to $\sR^d$ using a single fully-connected layer. To save on compute, during training we set the resolution to a maximum of $512\times512$ images, resizing on the long-edge of images. 
Additional implementation details 
% and inference efficiency analysis 
are in \cref{app:impl_details}.
Importantly, the use of a very lightweight CLIP encoder, combined with efficient Einsum matrix multiplications, ensure the inference overhead of \coolname is minimal. Indeed, \coolname inference takes $33.4$ seconds, compared to $32.4$ seconds for a single $r=128$ LoRA. See further details on our inference efficiency in \cref{app:overhead_analysis}.
We compare \coolname to four recent baselines: A standard \flux LoRA of similar parameter capacity (equivalent to \coolname with $N=1,r=128$), as well as three prior visual analogy methods based on Flux.1-Dev (RelationAdapter~\citep{gong2025relationadapter}, VisualCloze~\citep{li2025visualcloze} and EditTransfer~\citep{chen2025edit}).
We additionally compare to Diffusion Image Analogies (DIA)~\citep{vsubrtova2023diffusion} and PairEdit~\citep{lu2025pairedit}, in \cref{app:dia_comparison}. DIA relies on inversion into CLIP space and expensive per-sample backpropagation through the entire diffusion process of Stable Diffusion 1.4~\citep{rombach2021highresolutionLDM}. Nevertheless, for a fuller assessment, we compare their method with ours, including an additional variant we design to adapt DIA to Flux.1‑Kontext. PairEdit also uses per-sample optimization, training three Flux LoRAs for each input triplet: A concept LoRA and a content LoRAs trained in parallel to describe the transformation from $\rva$ to $\rva'$, and an inversion LoRA to reconstruct $\rvb$. Once all LoRAs are trained they can be used to generate $\rvb'$.

\subsubsection{Dataset.}
We train our model using the public Relation252k~\citep{gong2025relationadapter} set, which contains 16K analogy image pairs across 208 tasks. 
Since the train-set split of Relation252k is not fully publicly available, and only 10 unseen analogy tasks were released, we extend it with a custom validation set to evaluate visual analogies. Specifically, we focus on analogies that were not found in the training set, which we create in the following manner: First, we collect over 100 Unsplash\footnote{\url{https://unsplash.com/}} photos covering diverse concepts from three categories: animals, persons, and general objects.
Next, we create analogy pairs with a focus on two categories: transformations which are in-domain for the base text-to-image model, and transformations that are not. For in-domain transformations, we first use an LLM to summarize the training prompts for each task in the training-set of Relation252k, yielding $208$ representative prompts. Next, we ask the LLM to generate novel prompts that differ from the training set's prompts and manually verify that they match the given concept categories.
We filter prompts where \flux fails to produce a meaningful edit, and further randomly select 15 prompts per concept category from the remainder. We generate three images per prompt, obtaining a total of 135 analogy pairs. 
For out-of-domain analogies, we collect 18 community LoRAs for \flux from HuggingFace, which were trained to enable edits the base model failed with. We use these pre-trained LoRAs, and repeat the previous random sampling strategy to get 135 analogy pairs. 
Finally, we randomly select as the input images $\rvb$ two images from the matching concept category, with a similar aspect ratio to $\rva$ and $\rva'$, and crop them to the exact size.
Our resulting set contains $540$ analogy triplets across $90$ tasks and $3$ concept categories. Including the unseen set of Relation252K, this gives $100$ tasks across $840$ analogy triplets. On all experiments, we first aggregate the results per analogy task, and then aggregate over all tasks. 
More details appear in \cref{app:dataset_details}.

\subsection{Qualitative Evaluations}
Figures \ref{fig:teaser} and \ref{fig:our_res} include results of analogy-based editing using \coolname. Notably, the model generalizes to new tasks covering style transfer, background replacements, object insertion, object displacement and more. 
In \cref{fig:comparisons} we show qualitative comparisons of \coolname against the baselines. Notably, existing approaches either struggle with maintaining the content of the original image, or fail on some of the tasks. 
Crucially, some baselines struggle to maintain subject identities, \eg the cat in RelationAdapter or the woman in VisualCloze, or to accurately capture the details of the transformation, \eg the precise crown in the cat example. 
\coolname shows greater adaptability and succeeds in a wider range of tasks. Additional results appear in \cref{app:more_qual_results} and \cref{app:non-flux-gen}.

\subsection{Quantitative Evaluations}\label{sec:quantitative_evals}

\begin{figure*}[t]
    \centering
    \includegraphics[width=\linewidth]{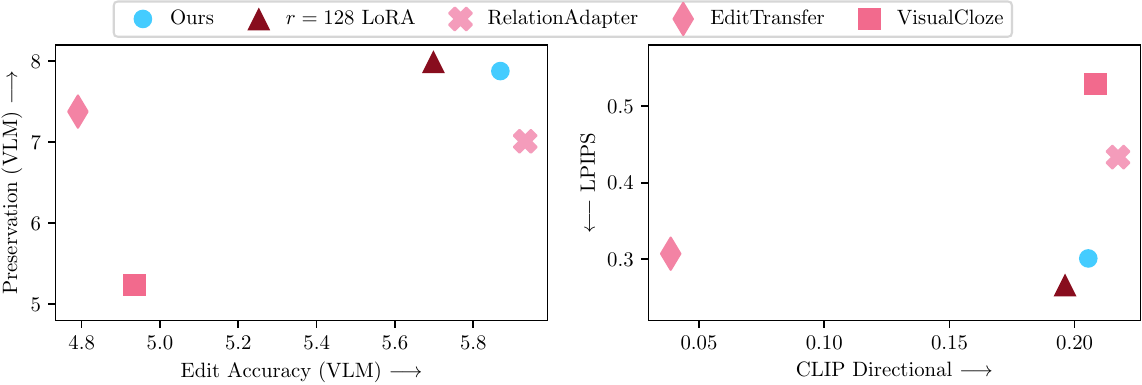}
    \caption{\textbf{Quantitative comparisons.} 
    (left) Accuracy of the applied edit and preservation of $\rvb$ in $\rvb'$ using Gemma-3~\citep{team2025gemma}. Top right is better. (right) CLIP directional similarity and LPIPS between $\rvb'$ and $\rvb$. Bottom-right is better. 
    Our method pushes the Pareto front of edit accuracy-preservation, achieving higher edit accuracy while strongly preserving the input image.
    }
    \label{fig:metrics}
\end{figure*}
\begin{figure*}[t]
    \centering
    \includegraphics[width=\linewidth]{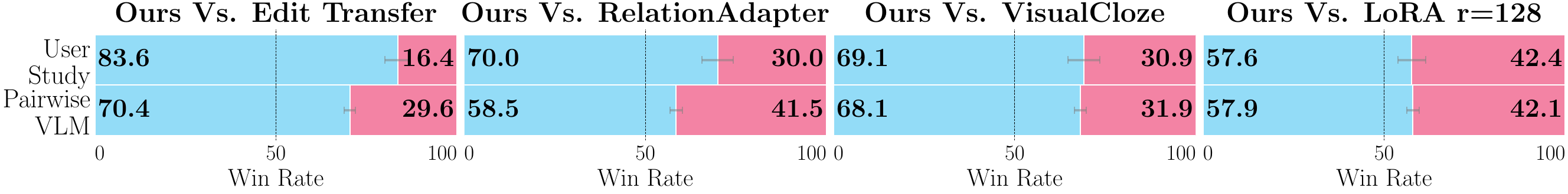}
    \caption{\textbf{Pairwise image comparisons.} 
    We compare \coolname to four baselines on overall edit quality preference via both a user study and using a  VLM. \coolname produces edits that are favored by both. Error bars are the $68\%$ Wilson score interval.
    }
    \label{fig:pairwise_vlm}
\end{figure*}

\subsubsection{Automated Evaluation Metrics.} For quantitative evaluations, we follow prior work~\citep{song2024lora,gu2024analogist,chen2025edit} and evaluate performance across standard metrics such as LPIPS~\citep{zhang2018unreasonable} between the source and generated image, and CLIP directional similarity between both analogy pairs. 
In addition, we build on recent image editing work~\citep{huang2025diffusion}, which demonstrates that VLMs often better correlate with human preference than CLIP-based methods, and implement a VLM-based assessment protocol. Specifically, we conduct two VLM-based experiments: In the first, we provide Gemma-3~\citep{team2025gemma} with $\{\rva, \rva', \rvb, \rvb'\}$, 
and ask the VLM to evaluate the quality of results on two criteria: consistency with the source image, and accuracy of the applied transformation relative to the reference transformation. We name these metrics \emph{Preservation (VLM)} and \emph{Edit Accuracy (VLM)}, respectively. 
As a second quality metric, 
we take a 2-alternative-forced-choice design (2AFC). We show Gemma-3 $\{\rva,\rva',\rvb\}$, the $\rvb'$ result of our model, and the $\rvb'$ result generated by a baseline, and ask it to select the image that best applies the analogy. We report this metric as \emph{Pairwise VLM}. The prompts given to the VLM and further details appear in \cref{app:vlm_eval_details}.
The results are shown in \cref{fig:metrics} and \cref{fig:pairwise_vlm}. When considering preservation and editing accuracy tradeoffs (\cref{fig:metrics}), our model pushes the Pareto front, achieving high edit accuracy while better maintaining the input's structure and appearance.

{
\begin{table*}[t]
\centering
    \caption{Results for the ablation study of \coolname described in \cref{sec:ablation}, for different hyperparameter and architecture choices. 
    }
    \label{tab:ablation_comprehensive}
    \scalebox{0.88}{
    \begin{tabular}{lcccc|cccc}
        \toprule
        \multirow{2}{*}{\textbf{Model}}  & \textbf{Pres. $\uparrow$}   & \textbf{Acc. $\uparrow$}  & \multirow{2}{*}{\textbf{LPIPS} $\downarrow$} & \textbf{CLIP} & \multicolumn{4}{c}{\textbf{Pairwise VLM (\%)} $\uparrow$} \\ 
        & \textbf{(VLM)} & \textbf{(VLM)} & & \textbf{Dir. $\uparrow$}  & \textbf{LoRA $r\!=\!128$} & \textbf{ET} & \textbf{VC} & \textbf{RA} \\ 
        \midrule
        \ourmethod{} (full, $r=4, N=32$) & 7.87 & 5.94 & 0.31 & 0.21 & 57.9 & 70.4 & 68.1 & 58.5  \\
        $+$ $r=16$ & 8.13 & 4.92 & 0.20 & 0.11 & 51.8 & 63.9 & 62.4 & 49.6 \\
        $+$ $r=16, N=8$ & 7.82 & 5.49 & 0.29 & 0.19 & 59.9 & 73.1 & 67.0 & 56.7 \\
        $+$ $N=16$ & 7.74 & 5.95 & 0.31 & 0.23 & 60.4 & 70.5 & 68.5 & 56.6 \\
        $+$ Tanh activation & 7.94 & 4.49 & 0.18 & 0.09 & 48.2 & 58.3 & 51.8 & 42.1 \\
        $+$ $2\times2$ Enc. Input & 7.90 & 5.75 & 0.28 & 0.20 & 61.9 & 73.3 & 68.2 & 53.9 \\
        $+$ SigLip2 & 7.83 & 5.82 & 0.31 & 0.21 & 59.0 & 71.7 & 71.5 & 55.5 \\
        $+$ SigLip2 \& $2\times2$ Enc. Input & 7.85 & 5.71 & 0.29 & 0.20 & 59.5 & 73.8 & 66.8 & 58.3 \\
        \bottomrule
    \end{tabular}
    }
\end{table*}
}

\subsubsection{User Study.}
Beyond automated metrics, we also conduct a two-alternative forced choice user study. We show each user a reference pair $(\rva, \rva')$, an input image $\rvb$, and two results (one from our model and one of a random baseline), in a randomized order, filtering out cases where no method succeeded in editing. Users are asked to select their preferred  editing result. In total, we collected responses from 33 users covering 45 image pairs. The results (\cref{fig:pairwise_vlm}) align with the automated metrics, showing that users favor our approach over all baselines.
In in \cref{app:vlm_eval_details} we additionally evaluate the alignment between the scores of the VLM and the preferences of humans.

All in all, our experiments demonstrate that our approach can meaningfully improve on the existing state of the art, and better generalize to unseen tasks.

\subsection{Ablations}\label{sec:ablation}

We next study the importance of different components of \coolname. 

\subsubsection{Capacity Effect.}
We compare \coolname across modified capacities in both basis sizes $N$ and ranks $r$. Specifically, we compare our original variation ($\{N\!=\!32,r\!=\!4\}$), 
with $\{N\!=\!8,r\!=\!16\}$, $\{N\!=\!16,r\!=\!4\}$ and $\{N\!=\!32,r\!=\!16\}$. 
We use the same evaluation setup as in \cref{sec:quantitative_evals}. Results are reported in \cref{tab:ablation_comprehensive}. Reducing the basis size while maintaining the capacity ($r\!=\!16,N\!=\!8$) leads to a slight drop in performance, as does simply reducing capacity ($r\!=\!4,N\!=\!16$). This highlights the importance of a large basis for generalization. 
Similarly, a \naive increase in rank can hamper editability, which we hypothesize to be a consequence of the data, leading to increased overfitting.
We provide additional capacity results for \coolname and a single LoRA with a higher capacity in \cref{app:more_quant_results}. Here, again, \naive parameter addition does not strictly correlate with better performance.

\subsubsection{Similarity Normalizing Function.}
The normalization function choice in \cref{eq:sim} can also affect the learned basis. For example, the used softmax is bound to $[0, 1]$, hence it cannot result in negative coefficients for any LoRA. An alternative approach is to use Tanh, which is instead bound to $[-1, 1]$.
In practice, we find it to drastically underperform. We propose that this may be due to Tanh allowing the model to compose mixed LoRAs with much greater norms, possibly taking the model too far out of domain. 
Another alternative can be a differential activation function, as proposed by several recent work~\cite{ye2025differential, misrahi2025difflora}.
However, we leave further investigation of activations to future work.

\subsubsection{Layout of Encoder Input.} In our approach, we elected to separately encode each of the conditioning analogy images using CLIP, and concatenate their representations. Our intuition is that CLIP requires resizing the image to $224\times224$, which can severely constrain the level of detail in each quadrant of the $2\times2$ grid that we provide Flux as a context. Moreover, concatenated features could allow the model to better understand which encoding represents each conditioning image (\ie $\rva$, $\rva'$ and $\rvb$), allowing it to better reason over the analogy. We verify this experimentally by comparing to a version that provides CLIP with just the context image (the $2\times2$ grid). 
As seen in \cref{tab:ablation_comprehensive}, this diminishes performance, mainly decreasing the editing-accuracy metrics.

\subsubsection{Alternative Image Encoders.} Although our approach uses CLIP~\citep{clip} as the encoder backbone, we validate our robustness to an alternative, common choice:  SigLIP2~\citep{tschannen2025siglip}. 
The results in \cref{tab:ablation_comprehensive} indicate that this change does not significantly alter our performances. We leave further tuning of encoders to future work.

\begin{figure*}[t]
    \centering
    \includegraphics[width=\linewidth
    ]{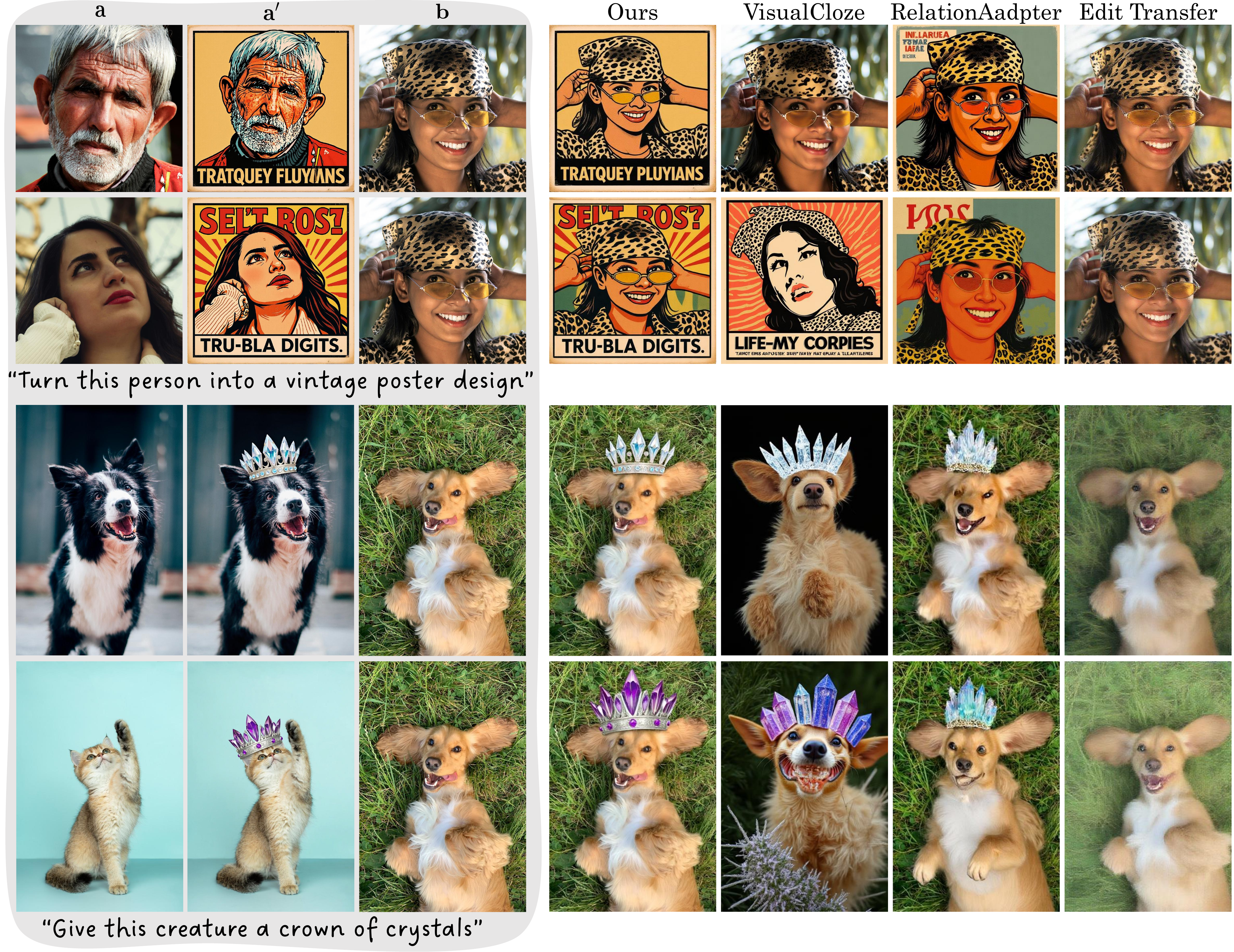}
    \caption{\textbf{Effect of different reference analogy pairs.} \coolname directly leverages the analogy pair to understand the details of the proposed task, applying an edit that is beyond just text-based editing based on the given prompt. For example, when the prompt is ``Give this create a crown of crystals'', the analogy context passes information on the amount and color of the crystals.}
    \label{fig:not_only_prompts}
\end{figure*}

\begin{figure}[th!]
    \centering
    \includegraphics[width=\linewidth]{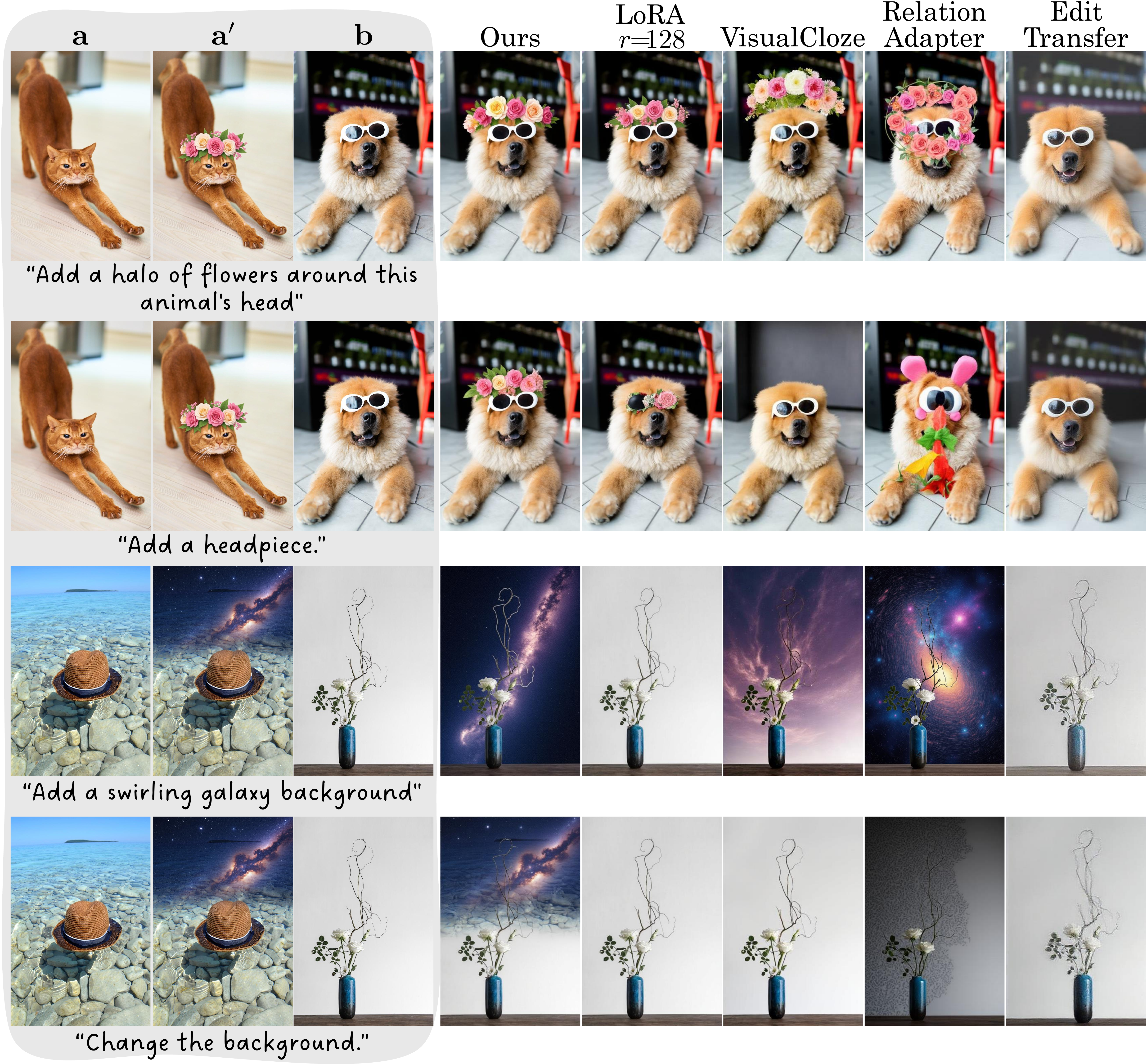}
    \caption{\textbf{Effect of less detailed prompts for the same input.}
    While we follow prior approaches in using textual prompts to understand the context of the task, the details of the applied transformation are achieved by understanding the reference pair of images $\rva$ and $\rva'$. For example, when asked to ``Add a halo of flowers around this animal's head'', most approaches add a similar-looking flower crown to the dogs head. However, when the prompt changes to only ``Add a headpiece'', only \coolname understands from $\rva:\rva'$ the headpiece is a flower crown that should be added on top of the dog's head.
    }
    \label{fig:varying_prompts_less_detail}
\end{figure}

\subsubsection{Importance of Prompts and Reference Images.}\label{sec:prompts_importance}

We follow existing baselines and use prompts to augment the model's understanding.
Since our goal is analogy based editing, and not simply text-based modification, we verify that indeed the output of our model depends not only on the text prompt, but on the analogy pair itself.
Specifically, we conduct two complementing experiments. First, we examine how the same input image, $\rvb$, reacts to different reference pairs $\{\rva,\rva'\}$ under the same editing prompt.
As can be seen in \cref{fig:not_only_prompts}, the reference pair dictates the details of the analogy task, and particularly the visual details that are not captured by the prompts.
For example, in the first row it copies the poster banner from the analogy image and adapts its specific style, and in the second row it matches the design and colors of the given crown.
In comparison, we observe that some of the baselines are insensitive to the analogy pair, instead relying almost entirely on the prompt. 
Here, again, it can be seen some of the baselines opt to change the identity in the image (\eg the dog of RelationAdapter and VisualCloze).
For the second experiment, we examine how the same input triplet $\{\rva,\rva',\rvb\}$ react to reducing the detail level in the prompt. Here, we use an LLM~\cite{gemini} to omit details from prompts in our evaluation set (instruction appears in \cref{app:vlm_eval_details}). 
When comparing the pairwise VLM preference rate for such detail-reduced prompts, our results are preferred over RelationAdapter, Edit-Transfer and VisualCloze 59.4\%, 70.4\% and 66.8\%. Qualitative examples can be seen in \cref{fig:varying_prompts_less_detail}. These results show that \coolname can better rely on the analogy images, when compared to prior work which relies heavily on the text prompt itself. As both experiments demonstrate, our approach has learned to perform analogy-based editing, and to a greater degree than the existing baselines. We experiment with unaligned prompts and input images in \cref{app:combinations}.

%% file: sec/5_discussion.tex
\section{Discussion}\label{sec:discussion}
We introduced \coolname, a modular framework for visual analogy completion that learns a basis of LoRA adapters and dynamically composes them using a shared encoder conditioned on the input analogy.
Our approach addresses the limitations of single-adapter fine-tuning or multi-adapter optimization at inference time by enabling flexible, layer-specific adaptations to diverse and unseen transformations. 
Through extensive comparisons, we showed how \coolname outperforms and generalizes better than competing naive LoRA-based methods across various visual analogy tasks. 
However, this generalization is not without limitations. For example, \coolname may still struggle with tasks that are significantly different from the training corpus (See \cref{app:limitations} for examples). 

Additionally, a strong assumption in image analogies is the access to a reference image pair $\{\rva, \rva'\}$ where the noticeable change depicts the transformation alone, while keeping the rest of the image details exactly the same (\eg depicting the same person in different poses in the same environment). In practical circumstances, finding such reference pairs in the wild, which convey the exact needed transformation, can be difficult.
Nevertheless, we show in \cref{app:different_a_atag} that similarity isn't strictly required, and 
\coolname can work on non-identical input pairs with textual guidance, within limit.
Indeed, an interesting future work could explore better decomposing the components of a depicted transformation (\eg the pose and the background), and allow for interactively choosing which components are used when applying the transformation to $\rvb$.

While our focus here is on visual analogy completion, a similar LoRA-basis approach could be broadly applicable, possibly replacing LoRAs in other tasks where generalization is needed. We hope to explore this direction in future work.

%% file: sec/app_implementation_details.tex
\section{Experimental Details}\label{app:implementation_details_sec}

\subsection{Implementation Details}\label{app:impl_details}

In all our experiments, we train for 10K steps on 1 H100 GPU, setting 8-bit AdamW~\citep{loshchilov2018decoupled} as the optimizer with a learning rate of $10^{-3}$, $\beta_1=0.9,\beta_2=0.99$, a weight decay value of $0.05$, and bfloat16 mixed-precision training.
We enable gradient checkpointing, and use a batch size of $6$ for all experiments, except for when $r=16,N=32$ where the batch size is set to $4$.
As for the encoders, the CLIP checkpoint we use is \texttt{openai/clip-vit-large-patch14}. For the SigLIP2 version in the ablations, we test \texttt{google/siglip2-base-patch16-224}. Both output a vector in $\sR^{768}$. 

\subsection{Efficiency Analysis}\label{app:overhead_analysis}
Compared to using a single, standard LoRA module, \coolname increases the computational load during inference due to using multiple LoRA modules.
However, the inference overhead of our approach is minimal. 
At inference time, \coolname requires passing the input triplet $\{\rva,\rva'\rvb\}$ through the CLIP encoder once and later combines the LoRA basis for each layer. 
As CLIP is very lightweight compared to \flux, and LoRA mixing is computed by an efficient Einsum matrix multiplication, this incurs negligible runtime overhead. 
Averaged over 100 images on an A100 GPU, \coolname inference takes $33.4\pm0.4$ seconds, compared to $32.4\pm0.3$ seconds for a single $r=128$ LoRA. This means only a $+3.1\%$ increase in runtime.
Furthermore, our current implementation re-computes the LoRA mixing every timestep. However, as the mixed LoRA of \coolname is fixed for all timesteps, a more efficient implementation of caching the mixing result from the first timestep can further reduce cost.

\subsection{Custom Inference Dataset} \label{app:dataset_details}

All images gathered from Unsplash for the inference dataset extension are free to use under the Unsplash license\footnote{\url{https://unsplash.com/license}}.
To simulate in-domain prompts, we use GPT-4o~\citep{hurst2024gpt} and Claude Sonnet 4~\citep{claude} to summarize the training prompts of Relation252k~\citep{gong2025relationadapter} as described in \cref{sec:experiments}, and generate novel prompts. The 15 randomly selected prompts per concept category (animals, objects, and persons) appear in \cref{tab:inference_set_generated_prompts}.
The 18 pre-trained LoRA adapters are sourced from HuggingFace\footnote{\url{https://https://huggingface.co/}}, and cover a range of transformation types such as style transfer, object modification, and artistic reinterpretation. Specifically, we use the following community LoRAs, with their provided trigger prompt:

\begin{itemize}%[leftmargin=0.5in]
    \item \href{https://huggingface.co/day-dream/MechAnything-Kontext-Dev-Lora}{\texttt{day-dream/MechAnything-Kontext-Dev-Lora}}
    \item \href{https://huggingface.co/drbaph/Fluffy-kontext-LoRA}{\texttt{drbaph/Fluffy-kontext-LoRA} }
    \item \href{https://huggingface.co/fal/3D-Game-Assets-Kontext-Dev-LoRA}{\texttt{fal/3D-Game-Assets-Kontext-Dev-LoRA} }
    \item \href{https://huggingface.co/fal/Cubist-Art-Kontext-Dev-LoRA}{\texttt{fal/Cubist-Art-Kontext-Dev-LoRA} }
    \item \href{https://huggingface.co/fal/Gouache-Art-Kontext-Dev-LoRA}{\texttt{fal/Gouache-Art-Kontext-Dev-LoRA} }
    \item \href{https://huggingface.co/fal/Minimalist-Art-Kontext-Dev-LoRA}{\texttt{fal/Minimalist-Art-Kontext-Dev-LoRA} }
    \item \href{https://huggingface.co/fal/Mosaic-Art-Kontext-Dev-LoRA}{\texttt{fal/Mosaic-Art-Kontext-Dev-LoRA} }
    \item \href{https://huggingface.co/fal/Pencil-Drawing-Kontext-Dev-LoRA}{\texttt{fal/Pencil-Drawing-Kontext-Dev-LoRA} }
    \item \href{https://huggingface.co/fal/Plushie-Kontext-Dev-LoRA}{\texttt{fal/Plushie-Kontext-Dev-LoRA}}
    \item \href{https://huggingface.co/fal/Pop-Art-Kontext-Dev-LoRA}{\texttt{fal/Pop-Art-Kontext-Dev-LoRA} }
    \item \href{https://huggingface.co/fal/Watercolor-Art-Kontext-Dev-LoRA}{\texttt{fal/Watercolor-Art-Kontext-Dev-LoRA} }
    \item \href{https://huggingface.co/gokaygokay/Bronze-Sculpture-Kontext-Dev-LoRA}{\texttt{gokaygokay/Bronze-Sculpture-Kontext-Dev-LoRA} }
    \item \href{https://huggingface.co/gokaygokay/Low-Poly-Kontext-Dev-LoRA}{\texttt{gokaygokay/Low-Poly-Kontext-Dev-LoRA} }
    \item \href{https://huggingface.co/gokaygokay/Marble-Sculpture-Kontext-Dev-LoRA}{\texttt{gokaygokay/Marble-Sculpture-Kontext-Dev-LoRA} }
    \item \href{https://huggingface.co/gokaygokay/Oil-Paint-Kontext-Dev-LoRA}{\texttt{gokaygokay/Oil-Paint-Kontext-Dev-LoRA} }
    \item \href{https://huggingface.co/Kontext-Style/Clay\_Toy\_lora}{\texttt{Kontext-Style/Clay\_Toy\_lora} }
    \item \href{https://huggingface.co/Kontext-Style/Ghibli\_lora}{\texttt{Kontext-Style/Ghibli\_lora} }
    \item \href{https://huggingface.co/Kontext-Style/Paper\_Cutting\_lora}{\texttt{Kontext-Style/Paper\_Cutting\_lora} }.
\end{itemize}

To match between $\rva,\rva'$ and $\rvb$ images of different sizes, we only choose $\rvb$ images with an original aspect ratio distanced 0.15 from the aspect ratio of $\rva$ and $\rva'$, and crop $\rvb$ to $\rva$'s aspect ratio. The images are resized to the same size with a maximum long edge of $512$ before entering \flux. 

\begin{table}[h]
    \caption{List of prompts generated for the inference sets}
    \centering
    \begin{tabularx}{\textwidth}{c|X}
         \textbf{Category} & \textbf{Prompt} \\
         \toprule
         Animals & Add a collar with a bell  \\
         Animals & Add a mountainous background \\
         Animals & Give this animal clockwork mechanical parts \\
         Animals & Add a flowing mane \\
         Animals & Add camouflage patterns \\
         Animals & Give this animal ethereal ghost-like transparency \\
         Animals & Add a flowing river background \\
         Animals & Add metallic golden fur highlights \\
         Animals & Give this animal translucent fairy wings \\
         Animals & Add a halo of fire \\
         Animals & Give this animal a fantastical set of armor \\
         Animals & Give this creature a crown of crystals \\
         Animals & Add a halo of flowers around this animal's head \\
         Animals & Give this animal bioluminescent markings \\
         Animals & Make this creature look sleepy \\
         Objects & Add a swirling galaxy background \\
         Objects & Render the object entirely as if it's made from hand-knitted or hand-crocheted yarn \\
         Objects & Add bioluminescent glowing elements \\
         Objects & Turn this into a candy or confectionery version \\
         Objects & Add flowing fabric or silk textures \\
         Objects & Turn this into a steampunk mechanical design \\
         Objects & Add intricate filigree patterns \\
         Objects & Turn this into a vintage advertisement poster \\
         Objects & Give this object a coat of rust \\
         Objects & Turn this photo into a cross-section diagram \\
         Objects & Make this look ancient and archaeological \\
         Objects & Turn this photo into a surrealist floating sculpture \\
         Objects & Make this look like it's growing moss \\
         Objects & Turn this photo into an architectural rendering \\
         Objects & Make this look like it's made of clouds \\
         Persons & Add a cape or cloak \\
         Persons & Add elaborate hairstyling with ornaments \\
         Persons & Make this person look heroic \\
         Persons & Add a serene, forested background \\
         Persons & Add golden hour lighting to this portrait \\
         Persons & Make this person look like a clown \\
         Persons & Add a swirling vortex background \\
         Persons & Add natural outdoor lighting to this portrait \\
         Persons & Make this person look like royalty \\
         Persons & Add body paint or decorative patterns \\
         Persons & Add temporary tattoos \\
         Persons & Turn this person into a holographic projection \\
         Persons & Add elaborate eye makeup \\
         Persons & Make this person look ethereal \\
         Persons & Turn this person into a steampunk portrait \\
         \bottomrule
    \end{tabularx}
    \label{tab:inference_set_generated_prompts}
\end{table}

\subsection{VLM Based Evaluation}\label{app:vlm_eval_details}

Part of our automated evaluation metrics include the use of Gemma-3~\citep{team2025gemma} as a VLM to evaluate our results.
We use two VLM-based experiments.
In the first, we ask the VLM to evaluate our results on two criteria: consistency with the source image $\rvb$ and accuracy of the applied transformation relative to the reference transformation described by $\{\rva,\rva'\}$. For this, we provide Gemma-3 with $\{\rva, \rva', \rvb, \rvb'\}$, and the following prompt:

\begin{lstlisting}
You are given 4 images: A (original image), A' (edited version of A), B (another original image), and B' (an output of an editing method). A, A' and B are reference images that are given to some editing method in order to generate B'. The method tries to infer the transformation that A underwent to produce A', and then tries (maybe unsuccessfully) to apply the exact same transformation to B - in order to generate B'. Your task is to evaluate the resulting B': Was the same transformation applied well?
Specifically, assess B' under two metrics, editing accuracy, and consistency with the original image B, 1-10 integers only: 
1) editing accuracy: Evaluate how closely B' applies the transformation seen from A to A'. Are there missing elements, are there redundant elements? Quantify the precision of the editing. 
2) consistency: Asses how well the edited image B' maintains the context of the original image B. Does it preserve the identity, objects, and layout in B that did not require a change, based on the infered transformation from A to A'? 
Consider in your evaluations other visual factors such as the localization of the edits, existence of redundant elements, style/strength/magnitude/colors of changes.
First, describe in detail what the transformation from A to A'. Then describe what elements of it are present or missing in B', detailing precisely what's wrong regarding each metric. 
Then, return a strict JSON with this scheme: 
\{"metrics":\{"accuracy":<1-10>,"consistency":<1-10>\}, 
"explanation":"the reasoning you described above"\}.
\end{lstlisting}

The JSON is parsed automatically, and we report the numeric values as \emph{Preservation (VLM)} and \emph{Edit Accuracy (VLM)}.

In the second quality metric, we take a 2-alternative-forced-choice design (2AFC). We show Gemma-3 five images: $\{\rva, \rva', \rvb\}$, the $\rvb'$ result of our model,
and the $\rvb'$ result generated by one baseline, and ask it to select the image that better applies the analogy via the following prompt:

% \texttt{
\begin{lstlisting}
You are given 5 images: A (original image), A' (edited version of A), B (another original image), and 2 B' images (outputs of 2 editing methods). A, A' and B are reference images that are given to some editing method in order to generate B'. The methods try to infer the transformation that A underwent to produce A', and then tries (maybe unsuccessfully) to apply the exact same transformation to B - in order to generate B'.
Your task is to evaluate the resulting B's: In which of the two methods was the same transformation applied well?
Specifically, assess B' under two metrics, editing accuracy, and consistency with the original image B, 1-10 integers only:
1) editing accuracy: Evaluate how closely B' applies the transformation seen from A to A'. Are there missing elements, are there redundant elements? Quantify the precision of the editing.
2) consistency: Asses how well the edited image B' maintains the context of the original image B. Does it preserve the identity, objects, and layout in B that did not require a change, based on the inferred transformation from A to A'?
Consider in your evaluations other visual factors such as the localization of the edits, existence of redundant elements, style/strength/magnitude/colors of changes.
First, describe in detail what the transformation from A to A'. Then describe what elements of it are present or missing in B'1 and B'2, detailing precisely what's wrong regarding each metric.
Then, return a strict JSON with this scheme: \{"better":<1 or 2>,"explanation":"the reasoning you described above"\}
\end{lstlisting}
% }

We report the winrates parsed from the JSON outputs as \emph{pairwise VLM}.

Additionally, we used an LLM~\citep{gemini} to modify the prompts for the varying prompts ablation reported in \cref{sec:ablation}. To generate these prompts, we used the following prompt:

\begin{lstlisting}
I am going to give you a file which contains 192 editing prompts.
Generate an output file, like that: 
For each prompt in the file, please edit it slightly, so the prompt will have less information of the specific editing task, but will still convey largely a similar editing task.
Please generate only 1 option per prompt in a file. 
The format of each output file should be a JSON like:
[{"original": original_prompt, "edit": edited_prompt},...].
Also, do not change the order of appearance, so that later when I check I can see which prompt turned into what by just comparing line numbers.
\end{lstlisting}

\subsubsection{Alignment With Humans.}
While VLMs have been used in the past as a metric aligned with human preference~\citep{huang2025diffusion,peng2025dreambench,ishikawa2025human}, even in the context of visual analogies~\citep{gong2025relationadapter}, we further validate their use in our task.
Specifically, we test the alignment between the scores of the VLM and the preferences of humans from our user study described in \cref{sec:experiments}. 
Following \citet{fu2023dreamsim}, 
We calculate the percentage of times the votes of each user agreed with the votes of the VLM and average over all users. We find this average user-VLM agreement to be $66.7\%$. As a baseline, we also compute the average agreement between different users. Namely, we compute the percentage of times the votes of each pair of users agreed and average over all user pairs. We find that this average user-user agreement is $74.2\%$. This means that our VLM based approach achieves a $89.9\%$ evaluation consistency with the evaluation of humans.
We also note that the mean standard deviation of user votes is $0.3423$, which is similar to the empirical standard deviation of the VLM's predictions from the users mean, which is given by $0.4649$.

%% file: sec/app_more_results.tex
\section{Additional Results}\label{app:more_results}

\subsection{Comparison to Diffusion Image Analogies (DIA) and PairEdit}\label{app:dia_comparison}

DIA~\citep{vsubrtova2023diffusion} is a per-sample optimization approach for image analogies, originally introduced for Stable Diffusion 1.4~\citep{rombach2021highresolutionLDM}. Specifically, this method uses backpropegation through the entire diffusion process to invert the reference and input images inot the models latent space, and its CLIP~\citep{clip} space. This takes 30GB for SD1.4 (and $>\!10[\text{mins}]$ per image on an H100 GPU). This is challenging with newer, larger models. 
Additionally, in contrast to our aproach, the reliance of DIA on inversion into CLIP space means it is not as adaptable to newer models (\eg, CLIP is not used in Flux.2).
Nevertheless, we compare to DIA to allow a more complete evaluation.

DIA is introduced as an interactive approach and presents 20 default hyperparameter options for the user to choose from. We therefore manually check all 20 options over 540 test images and perform the final evaluation with the best performing hyperparameter configuration ($\sigma\!=\!12$, $\lambda\!=\!0.63$). 
Finally, we also adapt DIA to Flux.1-Kontext.
First, we replace their $\rvb$ inversion by setting $\rvb$ as the context image. Next, as backpropagation through the entire flow process is computationally infeasible, we replace the CLIP embedding inversion by setting the CLIP ($\gE$) embeddings as $\smash{\gE(\rvb)\!+\!\gE(\rva')\!-\!\gE(\rva)}$.
Finally, Flux.1-Kontext includes an additional text encoder, T5~\cite{2020t5}. Therefore, for a fairer comparison, we provide the editing prompt to the T5 encoder. We term this approach \emph{DIA-Kontext}.

PairEdit~\citep{lu2025pairedit} is also a per-sample optimization approach for examplar-based image-editing, designed for Flux.1-Dev. Specifically, this method trains three distinct LoRAs for each input triplet $\{\rva,\rva'\rvb\}$. Given $\rva$ and $\rva'$, the method first jointly trains a content LoRA, which reconstructs the source image using the standard flow loss, and a semantic LoRA, which aims to capture the semantic transformation between the two images by optimizing a semantic loss. The join training encourages the semantic LoRA to disentangle the semantic differences from the image content.
Next, to apply this transformation to a new image $\rvb$, PairEdit requires training an inversion LoRA of $\rvb$. Finally, at inference the inversion LoRA and the semantic LoRA are aplpied together to produce $\rvb'$. This optimization approach takes dozens of minutes and intensive compute to edit a single new image (more than 25 minutes on an A100 GPU). In contrast, \coolname requires no test time training.

\begin{figure}[ht]
    \centering
    \includegraphics[width=\linewidth]{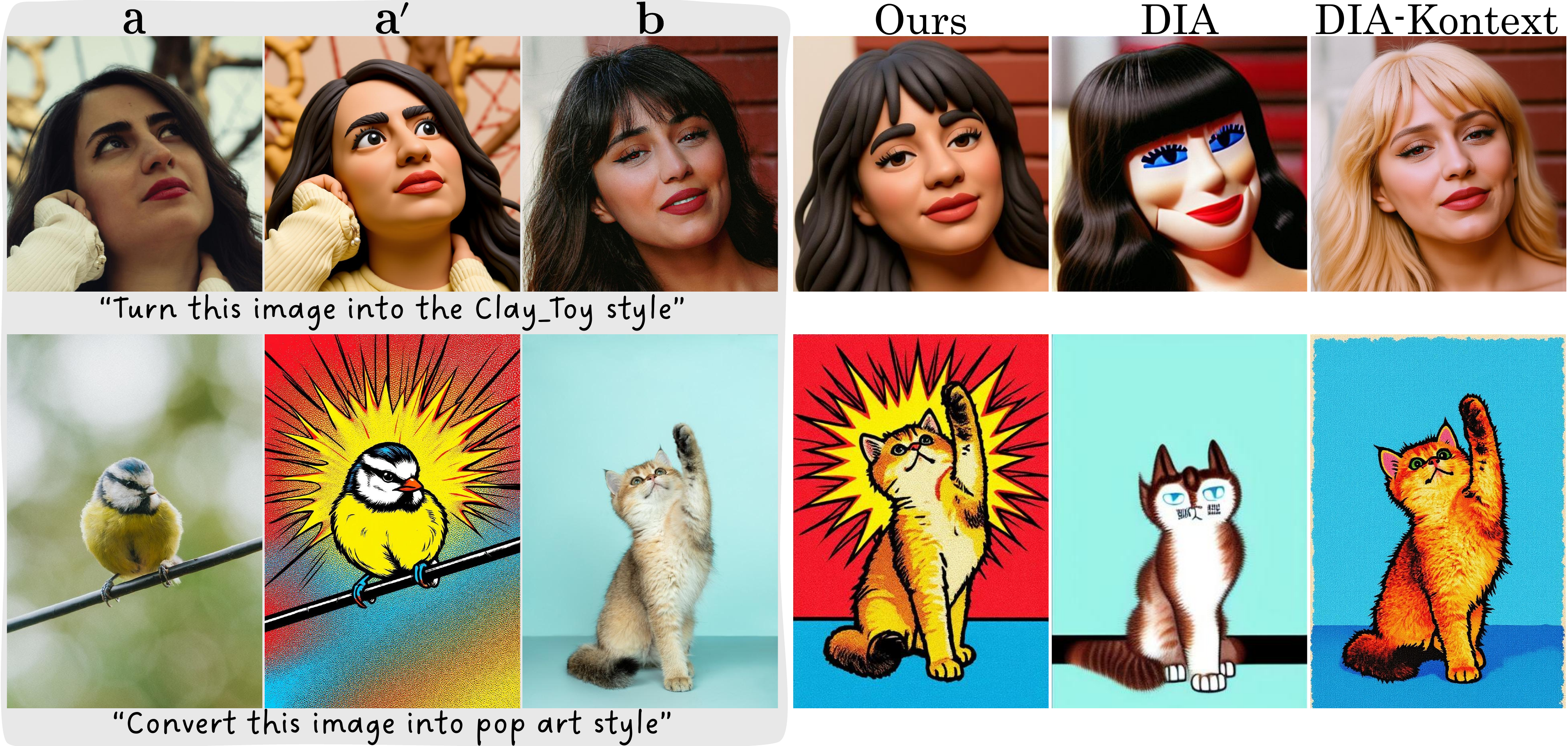}
    \caption{\textbf{Qualitative comparison with DIA \& DIA-Kontext.} \coolname outperforms DIA and our Flux-adapter variant DIA-Kontext by a large margin. Notably, DIA-Kontext does preserves the identity of both the cat and the woman, yet struggles to accurately perform the transformation.}
    \label{fig:dia}
\end{figure}

\begin{figure}[ht!]
    \centering
    \includegraphics[width=\linewidth]{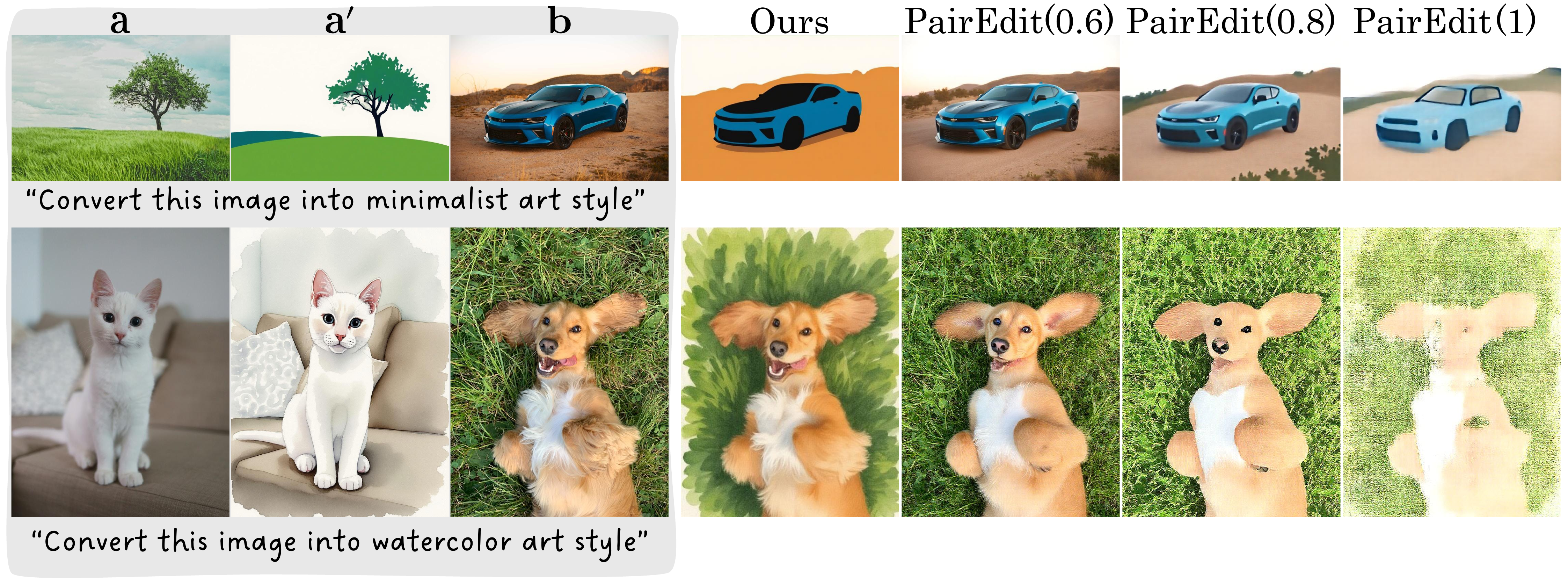}
    \caption{\textbf{Qualitative comparison with PairEdit.} \coolname outperforms PairEdit, across multiple PairEdit strengths variations.}
    \label{fig:pairedit}
\end{figure}

Qualitative results can be seen in \cref{fig:dia,fig:pairedit}, and quantitative evluations appear in \cref{tab:dia}. As can be seen, \coolname outperforms all approaches. Specifically, while this performance boost may seem trivial over DIA due to their use of a weaker pre-trained model (SD1.4), \coolname also outperforms DIA-Kontext. DIA-Kontext preserves the identity of the object, yet struggles to accurately perform the transformation. While PairEdit can sometimes score lower on LPIPS, its overall edit-balance of edit-adherence and content preservation is lacking, yielding sub-par results across all metrics.

\begin{table}[ht!]
    \centering
    \setlength{\tabcolsep}{4pt}
    \caption{Quantitave comparison with DIA, DIA-Kontext, and PairEdit}
    \begin{tabular}{lcccc|c}
        \toprule
        \multirow{2}{*}{\textbf{Method}}  & \textbf{Pres. $\uparrow$}   & \textbf{Acc. $\uparrow$}  & \multirow{2}{*}{\textbf{LPIPS} $\downarrow$} & \textbf{CLIP} & \textbf{\underline{Our} Pairwise $\uparrow$} \\ 
        & \textbf{(VLM)} & \textbf{(VLM)} & & \textbf{Dir. $\uparrow$}  &
        \textbf{VLM vs.~(\%)} \\ 
        \midrule
        \textbf{Ours} & 7.87 & 5.94 & 0.31 & 0.21 & -- \\
        DIA & 3.56 & 3.44 & 0.59 & 0.12 & $86.5\%$ \\
        DIA-Kontext & 4.78 & 3.56 & 0.63 & 0.17 & $68.6\%$ \\
        PairEdit $s=0.6$ & 6.68 & 4.55 & 0.24 & 0.11 & $75.6\%$ \\
        PairEdit $s=0.8$ & 4.44 & 3.60 & 0.44 & 0.16 & $86.3\%$ \\
        \bottomrule
    \end{tabular}
    \label{tab:dia}
\end{table}

\clearpage

\subsection{Additional Quantitative Results}\label{app:more_quant_results}

We conduct two additional experiments with \coolname of a larger capacity ($r=4,N=64$), as well as a single LoRA with higher capacity, of $r=256$. The results, along with a detailed table of the numerical values in \cref{fig:metrics}, appear in \cref{tab:ablation_app}.
As evident, \naive parameter addition does not strictly correlate with better performance, and can cause the methods to overfit.

{
\begin{table*}[h]
    \centering
    \caption{additional results for the ablation study of \coolname described in \cref{sec:ablation}, for different hyperparameter and architecture choices. 
    }
    \label{tab:ablation_app}
    % \footnotesize
    \scalebox{0.86}{
    \begin{tabular}{lcccc|cccc}
        \toprule
        \multirow{2}{*}{\textbf{Model}}  & \textbf{Pres. $\uparrow$}   & \textbf{Acc. $\uparrow$}  & \multirow{2}{*}{\textbf{LPIPS} $\downarrow$} & \textbf{CLIP} & \multicolumn{4}{c}{\textbf{Pairwise VLM (\%)} $\uparrow$} \\ 
        & \textbf{(VLM)} & \textbf{(VLM)} & & \textbf{Dir. $\uparrow$}  & \textbf{LoRA $r=128$} & \textbf{ET} & \textbf{VC} & \textbf{RA} \\ 
        \midrule
        \coolname (full, $r=4, N=32$) & 7.87 & 5.94 & 0.31 & 0.21 & 57.9 & 70.4 & 68.1 & 58.5  \\
            \coolname on $(r=4,N=64)$ & 7.80 & 5.48 & 0.27 & 0.19 & 56.5 & 67.7 & 66.3 & 52.6 \\
            LoRA $r=128$       & 7.99 & 5.70 & 0.27 & 0.20 & N/A & N/A & N/A & N/A \\
            LoRA $r=256$       & 7.88 & 5.48 & 0.26 & 0.18 & N/A & N/A & N/A & N/A \\
            VisualCloze       & 5.24 & 4.93 & 0.53 & 0.21 & N/A & N/A & N/A & N/A \\
            RelationAdapter       & 7.01 & 5.93 & 0.43 & 0.22 & N/A & N/A & N/A & N/A \\
            Edit-Transfer       & 7.38 & 4.79 & 0.31 & 0.04 & N/A & N/A & N/A & N/A \\
        \bottomrule
    \end{tabular}
    }
\end{table*}
}

\subsection{Additional Qualitative Results}\label{app:more_qual_results}
\begin{figure}[th]
    \centering
    \includegraphics[width=\linewidth]{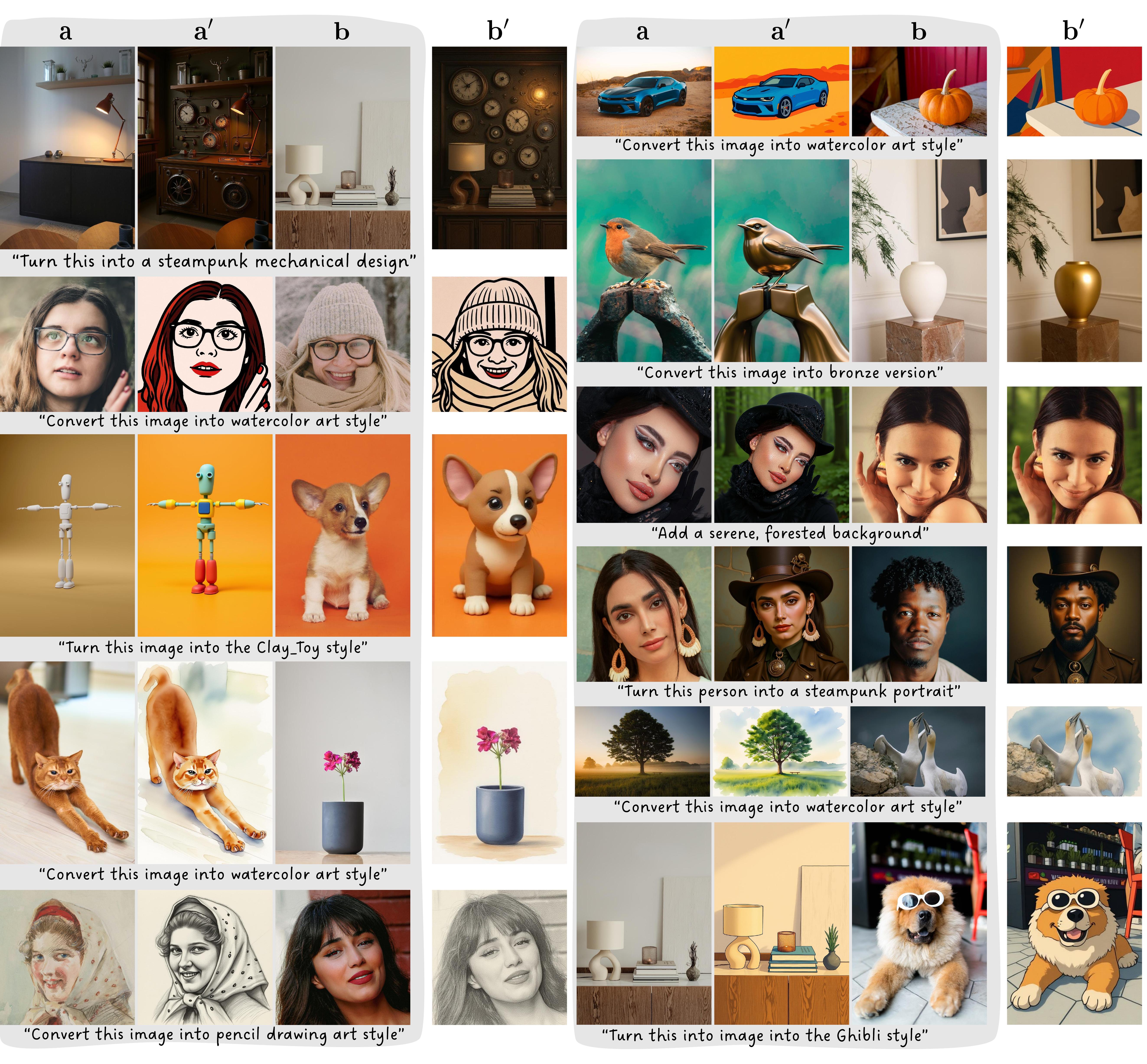}
    \caption{\textbf{\coolname visual analoy results.} The use of a LoRA Basis allows LoRBA to generalize to a wide varity of new analogy tasks, from changing given images to certain styles such as clay toys or bronze sculptures, changing the backgrounds, or changing the cloths of the person. Please zoom in for more details.}
    \label{fig:app_more_results}
\end{figure}

\begin{figure}[t]
    \centering
    \includegraphics[width=\linewidth]{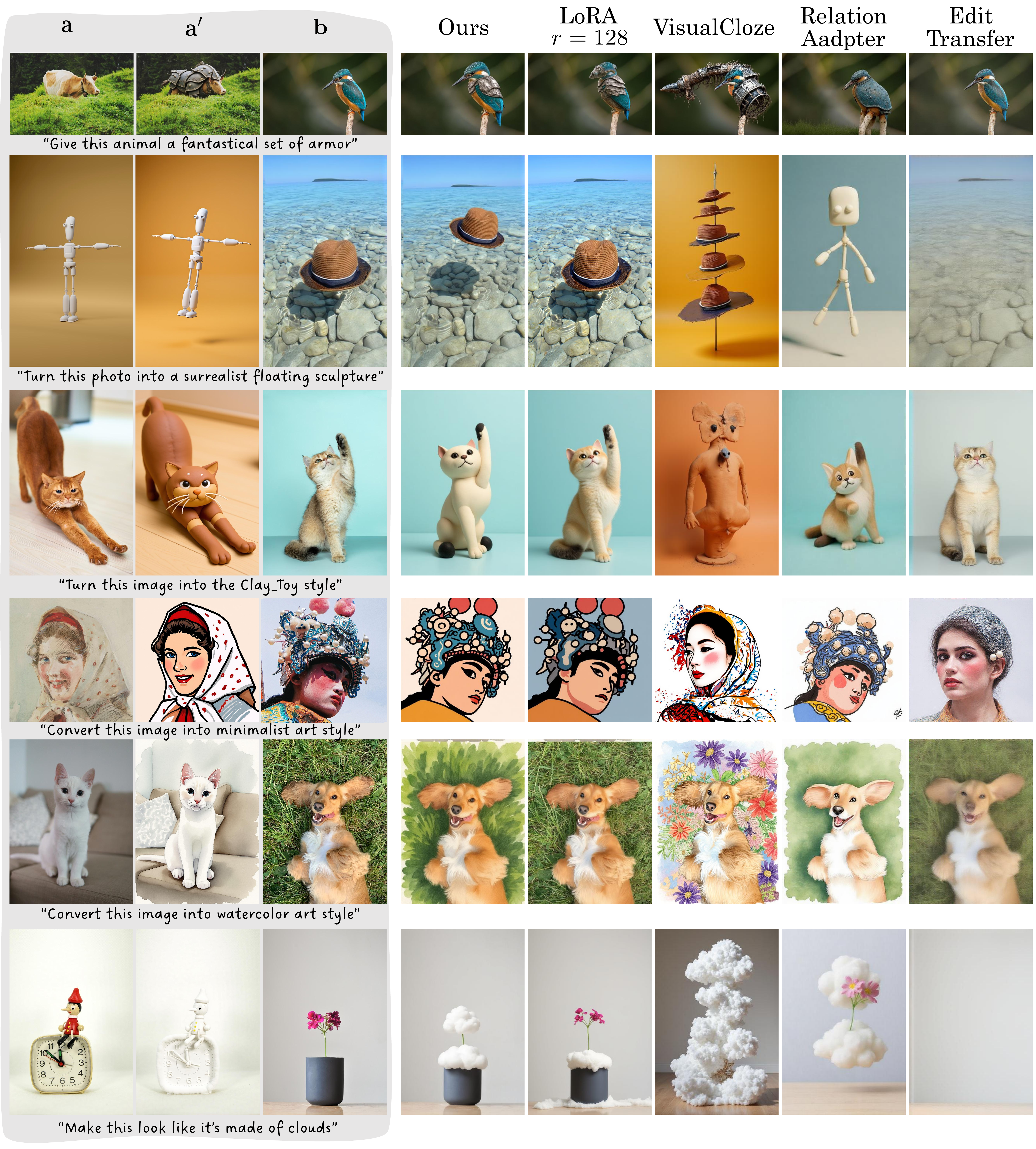}
    \caption{\textbf{Comparisons with baseline methods on unseen tasks}. Our approach generalizes more across diverse tasks, and better maintains the visual details of both the subject and the analogy.}
    \label{fig:app_more_comp}
\end{figure}

We provide additional qualitative results of our method in \cref{fig:app_more_results}, as well as more comparisons of our method to the 4 baselines from \cref{sec:experiments} in \cref{fig:app_more_comp}.

\clearpage

\subsection{Sensitivity to Non-Identical Input Pairs}\label{app:different_a_atag}

\begin{figure}[t]
    \centering
    \includegraphics[width=0.7\linewidth]{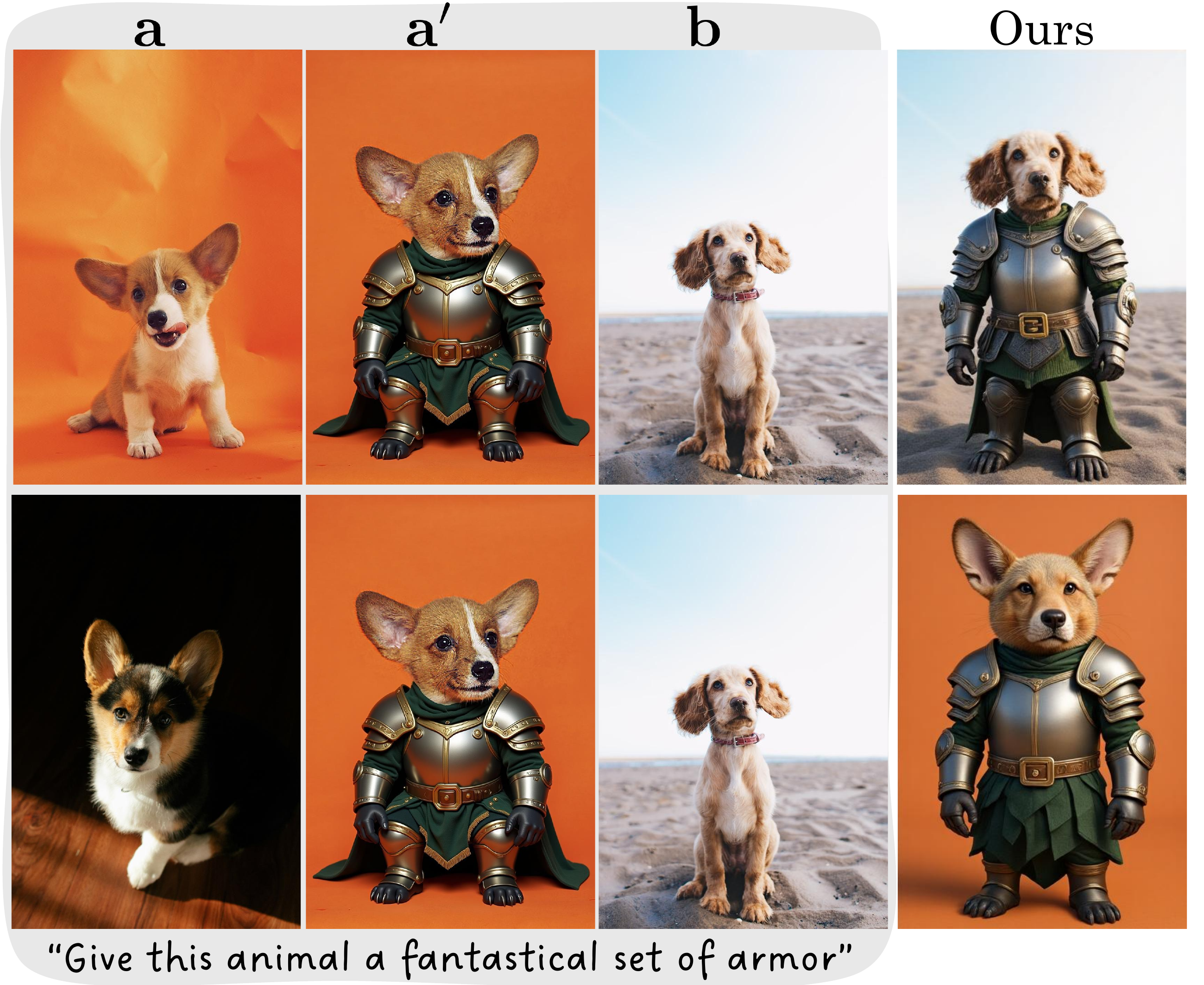}
    \caption{\textbf{The effect of using non-identical $\rva:\rva'$.} Here, we use the top-right example from \cref{fig:teaser}. While exact alignment of $\rva$ and $\rva'$ helps, and \coolname was trained on such samples, empirically we find \coolname can work on images depicting more than the intended transformation, with text guidance, to some limit.}

    \label{fig:different_a_atag}
\end{figure}

As mentioned in \cref{sec:discussion}, in image analogies a commonplace assumption is the access to a reference image pair $\{\rva, \rva'\}$ where the noticeable change depicts the transformation alone, while keeping the rest of the image details exactly the same.
However, in practical circumstances, finding such reference pair of input images, which depict only a single modification, is hard to come by in the wild. Indeed, such pairs are more common when shooting a series of images by design, \eg in a professional photo-shoot. 
Still, we empirically find that \coolname can work on non-identical input pairs with textual guidance, within limit.
For example, in \cref{fig:different_a_atag} we repeat the example from \cref{fig:teaser} with different $\rva$ images. As can be seen, changing the position of the dog in $\rva$ did not negatively effect the resulting $\rvb$, as the edit is accurate and the identity of the dog in $\rvb$ is preserved. However, changing the $\rva$ to a different dog distorts the output $\rvb$, leading to loss of identity preservation, as well as the image background.

\clearpage

\subsection{Effect of Misalignment in Text and Image Inputs}\label{app:combinations}

\begin{figure}[t]
    \centering
    \includegraphics[width=0.7\linewidth]{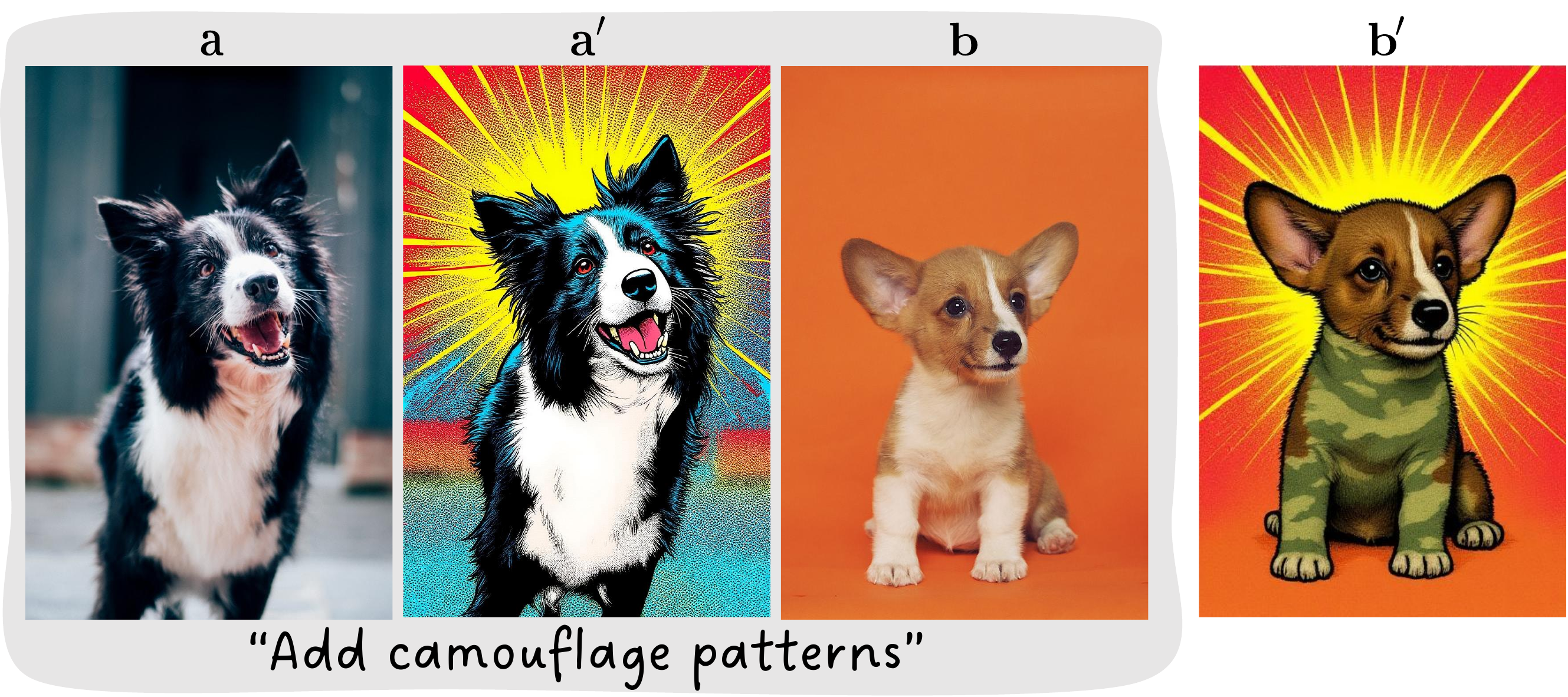}
    \caption{\textbf{The effect of using unaligned textual prompt $c$ and input images pair $\rva:\rva'$.} 
    Using text prompts that describe a different edit than that seen in the analogy images pair can produce a combined-editing effect, displaying elements from both types of input.
    }
    \label{fig:combinations}
\end{figure}

As discussed in \cref{sec:prompts_importance}, while we follow existing baselines in using prompts, this necessitates validating the balance of prompt and input images effect on the output. 
In \cref{sec:prompts_importance} we conduct two experiments to provide insight on the effects of different prompts on the output, and the effect of different image inputs on the output.
Here, we further investigate the effect of the two inputs by examining how misalignment affects the output. Specifically, we provide to \coolname a conditioning prompt $c$ which does not describe the analogy seen in the set a reference image pairs $\{\rva,\rva'\}$.
Indeed, \coolname was not trained on such inputs, and was trained on aligned data, where the prompts roughly describe the analogy seen in the input pair. 
Empirically, we find this can either cause \coolname to ignore one of the inputs, or create an interesting combined-editing effect.
As seen in \cref{fig:combinations}, this combined-editing effect results an output image which displays features seen in the analogy pair as well as features understood from the textual prompt. This further strengthens that \coolname has learned to perform textually-guided analogy-based editing, using both the input images and the textual prompt.

\clearpage

\subsection{Sensitivity to Non-Flux Generated Images}\label{app:non-flux-gen}

\begin{figure}
    \centering
    \includegraphics[width=\linewidth]{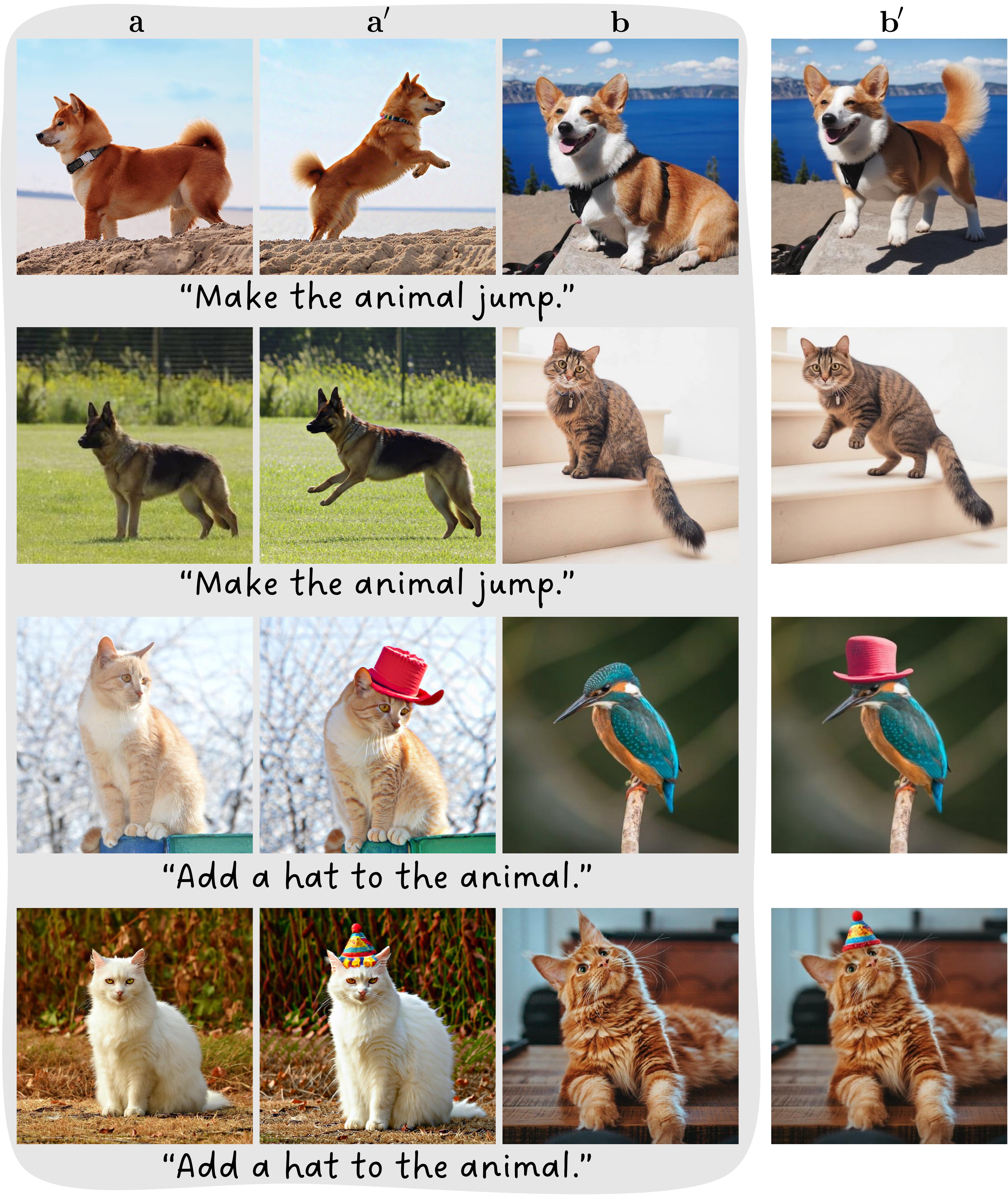}
    \caption{\textbf{Sensitivity to reference image pairs not generated by Flux.}
    We explicitly test \coolname on cases where the input reference pair of images, $\{\rva,\rva'\}$, was not generated by Flux. Specifically, here the images were drawn from TEdBench, where $\rva'$ was generated by Imagic~\citep{kawar2023imagic}. The use of image not generated by Flux does not hamper the performance of \coolname, verifying it is not biased towards Flux-generated images.
    }
    \label{fig:imagic}
\end{figure}

Our evaluation set is composed of the test set of Relation252K~\citep{gong2025relationadapter} as well as our custom set of edited Unsplash images.
The reference image pairs in Relation252K were edited by using MidJourney, or curated from existing benchmark datasets. 
Our custom set was generated by instructing Flux with editing prompts. 
In \cref{fig:imagic}, we further verify the sensitivity of \coolname to reference image pairs not generated by Flux. Specifically, we test over images from TEdBench, which contains images edited by Imagic~\citep{kawar2023imagic }. As can be seen, the use of non-Flux generated images as the reference pair does not hampers the performance of \coolname.

\clearpage

\subsection{Failure Cases}\label{app:limitations}
\begin{figure}[h]
    \centering
    \includegraphics[width=\linewidth]{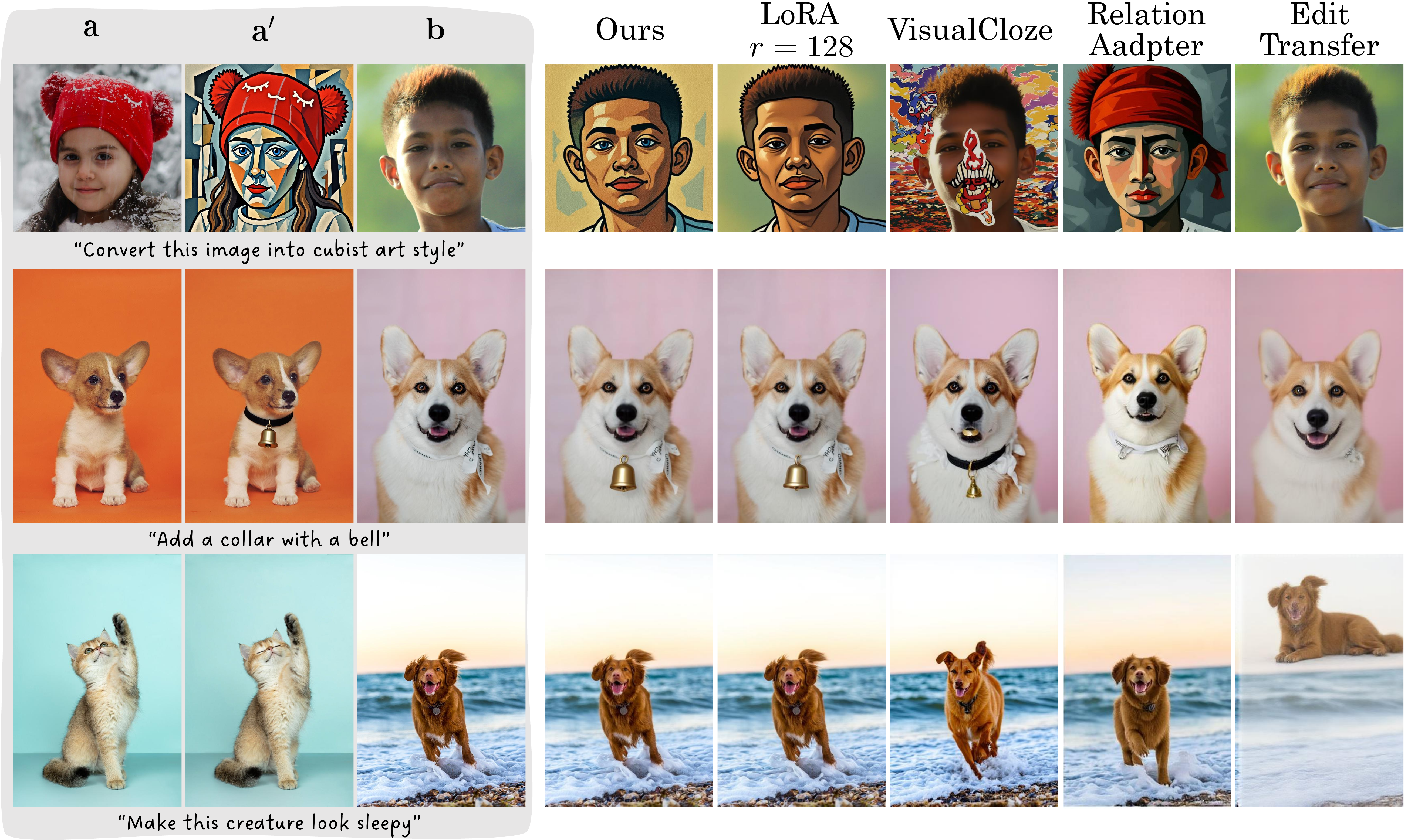}
    \caption{\textbf{Examples of failure cases of \coolname.} When the tasks are significantly different from those seen in the training corpus (first row) or require modifying only small details (third row) \coolname might struggle with applying the needed transformation. Additionally, \coolname might only partly apply a transformation if it requires modifying multiple elements (second row).}
    \label{fig:limitations}
\end{figure}

\coolname better generalizes than competing approaches across various visual analogy tasks. 
However, this generalization is not without limitations. Here we present some failure cases of \coolname. 
Specifically, \coolname may still struggle with tasks that are significantly different from the training corpus. 
As can be seen in the first row of \cref{fig:limitations}, turning $\rvb$ into a cubist art style proved difficult for all approaches. Here, $\rva'$ was created by using a community LoRA, such that the base model itself also had difficulties in creating this specific style.
Additionally, as seen in the second row of \cref{fig:limitations}, \coolname might only partly understand the analogy if it includes several components.
Here, the analogical difference includes a black collar and a large bell, yet \coolname only added the bell. 
Finally, if the difference between the reference pair $\rva$ and $\rva'$ is small, such as closing the eyes of the animal (third row of \cref{fig:limitations}), \coolname might struggle reflecting the changes. This might be due to the use of a CLIP models which downsamples the images to a size of $224\times224$, which might make it less sensitive to small details.

